\documentclass{article} % For LaTeX2e
\usepackage[final]{colm2025_conference}

\usepackage{microtype}
\usepackage{hyperref}
\usepackage{url}
\usepackage{booktabs}

\usepackage{lineno}

\definecolor{darkblue}{rgb}{0, 0, 0.5}
\hypersetup{colorlinks=true, citecolor=darkblue, linkcolor=darkblue, urlcolor=darkblue}

\usepackage{amsmath}
\usepackage{amssymb}
\usepackage{mathtools}
\usepackage{amsthm}
\usepackage{xspace}

\usepackage{hyperref}       % hyperlinks
\usepackage{url}            % simple URL typesetting
\usepackage{booktabs}       % professional-quality tables
\usepackage{amsfonts}       % blackboard math symbols
\usepackage{graphicx}
\usepackage{multirow}
\usepackage{lipsum}
\usepackage{indentfirst}
\usepackage{dsfont}
\usepackage{relsize}
\usepackage{svg}
\usepackage{amsthm}
\usepackage{xspace}
\usepackage{caption}
\usepackage{subcaption}
\usepackage{listings}
\usepackage{threeparttable}
\usepackage{array}
\usepackage{float}
\usepackage{ragged2e}
\usepackage{xcolor}
\usepackage{amssymb}
\usepackage{tabularx}
\usepackage{tikz}
\usepackage{bbm}
\usepackage{enumitem}
\usepackage{xurl}
\usepackage[most]{tcolorbox}
\tcbuselibrary{theorems}
\usepackage{cleveref}

\newtcbtheorem[]{exmp}{System Prompt}%
{fonttitle=\bfseries, left=.02in, right=.02in,bottom=.02in, top=.02in}{exmp}

\RequirePackage{algorithm}
\RequirePackage{algorithmic}

\usepackage{pifont}
\newcommand{\cmark}{\textcolor{green}{\ding{51}}}%
\newcommand{\xmark}{\textcolor{red}{\ding{55}}}%

\usepackage{comment}
\usepackage{xspace}

\newcommand{\eg}{\textit{e.g.},\xspace}
\newcommand{\ie}{\textit{i.e.},\xspace}
\DeclareMathOperator*{\argmax}{arg\,max}

\renewcommand{\paragraph}[1]{
\noindent \textbf{#1.~} 
 }
\setenumerate{itemsep=0pt,leftmargin=0.1in}
\setitemize{itemsep=0pt,leftmargin=0.1in}

\title{Resource-efficient Inference with Foundation Model\smallskip\\ Programs}

\author{Lunyiu Nie$^{1}$ \quad Zhimin Ding$^{2}$ \quad Kevin Yu$^{1}$ \quad Marco Cheung$^{1}$ \\ \textbf{Christopher Jermaine}$^{2}$ \quad \textbf{Swarat Chaudhuri}$^{1}$ \\
$^1$The University of Texas at Austin \quad
$^2$Rice University\\
\href{mailto:lynie@utexas.edu}{\texttt{lynie@utexas.edu}}, \href{mailto:lynie@utexas.edu}{\texttt{swarat@cs.utexas.edu}}
}

\newcommand{\revise}[1]{\textcolor{black}{#1}}

\begin{document}

\ifcolmsubmission
\linenumbers
\fi

\maketitle

\begin{abstract}

The inference-time resource costs of large language and vision models present a growing challenge in the deployment of agentic systems. 
We propose the use of \emph{foundation model programs}, i.e., programs that can invoke foundation models with varying resource costs and performance, as an approach to this problem. Specifically, we present a method that translates a task into a program, then learns a policy for resource allocation that, on each input, selects foundation model ``backends" for each program module. The policy uses smaller, cheaper backends to handle simpler subtasks, while allowing more complex subtasks to leverage larger, more capable models. We evaluate the method on two new ``streaming" visual question-answering tasks in which a system answers a question on a sequence of inputs, receiving ground-truth feedback after each answer. Compared to monolithic multi-modal models, our implementation achieves up to 98\% resource savings with minimal accuracy loss, demonstrating its potential for scalable and resource-efficient multi-modal inference \footnote{Source code and benchmarks are available at 
% \url{https://anonymous.4open.science/r/FMP}.}. 
\url{https://github.com/Flitternie/FMProgramming}.}. 
\end{abstract}

% \begin{abstract}

% The inference costs of large language and vision models present a growing challenge in production deployments, especially under streaming or real-time constraints. 
% We propose \emph{Foundation Model Programming} (FMP), a general framework for cost-efficient inference in complex, multi-modal tasks. By exploiting task structure and tailoring submodules to each input's complexity, FMP can trace an optimal tradeoff curve on the Pareto frontier of resource usage versus performance.
% FMP includes an \emph{offline code generation} step that decomposes a complex task into submodules, followed by an \emph{online resource allocation} policy that dynamically selects FM backends for each subtask via a structured REINFORCE algorithm and gradient-based Thompson Sampling.
% Empirical results show that FMP achieves up to 98\% resource savings with minimal accuracy loss on newly introduced streaming benchmarks for visual question answering, demonstrating its potential for scalable and resource-efficient multi-modal inference \footnote{The source code and the benchmarks are available at \url{https://anonymous.4open.science/r/FMP}.}. 
% \end{abstract}

\newcommand{\name}{\textsc{FMP}\xspace}
\section{Introduction}

Foundation models (FMs) have reshaped the landscape of machine learning over the past few years, demonstrating unprecedented capabilities in language understanding \citep{achiam2023gpt, dubey2024llama}, complex reasoning \citep{lu2024chameleon, gupta2024language}, and multimodal tasks \citep{li2022blip, liu2024improved}.
While much of the community's attention has focused on their training costs, the \emph{inference-time resource use} of FMs is increasingly becoming a practical bottleneck. For commercial applications that require real-time responses --- for instance, continuous streams of user queries to a multimodal large language model (MLLM) --- computational overhead and high latency can severely degrade user experience and inflate operational expenses \cite{xu2024survey}.

% A popular strategy to mitigate FM inference overhead is to adopt routing or cascading mechanisms \citep{chen2023frugalgpt, shnitzer2023large, lu2023routing, nie2024online}, whereby an input is either routed to a smaller model if deemed ``easy'' or escalated to a larger model if deemed ``difficult.''  While these methods can be effective, they typically overlook the compositional structure underlying many tasks. For example, a visual question answering (VQA) system that receives the query ``Is there a cat sitting or laying on a laptop keyboard in the image?'' might first call a small, inexpensive object detection model to check whether both a cat and a laptop are present, and only if that condition is met would it invoke a more powerful vision-language model (VLM) for finer-grained reasoning. Current approaches ignore this task structure, instead adopting a one-size-fits-all architecture, leaving substantial efficiency gains on the table.

In this paper, we propose the use of \emph{foundation model programs} (FMPs) --- code in Python-like languages that can call into a variety of specialized vision and language models as subroutines --- as a way to address this problem. Such programs have been previously motivated by the interpretability and composability they bring to multi-step tasks \citep{suris2023vipergpt,gupta2023visual,subramanian2023modular} and agentic AI systems \citep{wu2023autogen, khattab2023dspy}. Our insight is that they can also 
%Unlike end-to-end MLLMs that treat each query as a single monolithic computation, foundational model approaches \cite{suris2023vipergpt,chaudhuri2021neurosymbolic} decompose tasks into symbolic modules or subprograms, each of which invokes specialized FMs. 
%This modular design not only brings interpretability and flexibility to multi-step tasks but also 
enable fine-grained decisions about resource allocation: \textbf{simpler subtasks can rely on smaller, cheaper backends while more complex components can leverage larger, more capable models}.  

\begin{figure*}[t]
     \centering
     \includegraphics[width=\textwidth]{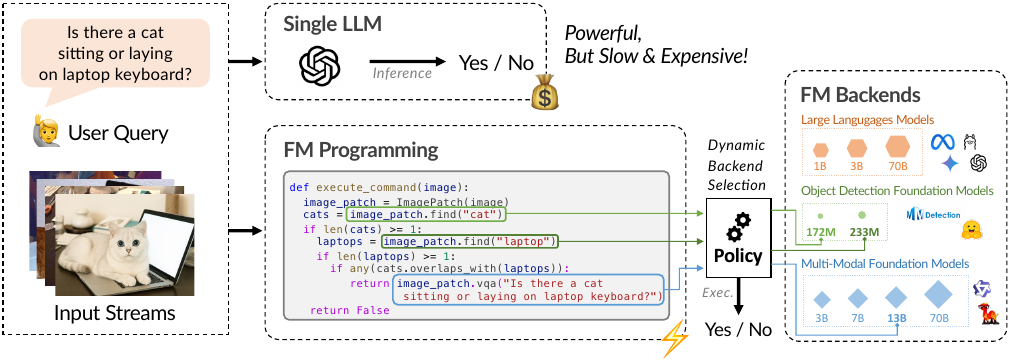}
     \caption{Illustration of an FM program synthesizing a VQA task by decomposing the task into sub-components. At runtime, the resource-efficient FM programming framework dynamically selects FM backends based on the task and input complexity to optimize accuracy and resource efficiency in real-time processing.} 
     \label{fig:background}
     \vspace{-0.15in}
\end{figure*}

%By adapting to each input's complexity, our method can traverse the Pareto frontier of resource usage vs.\ performance, preserving accuracy on demanding subtasks while saving costs for simpler ones.
Concretely, we propose a framework of \emph{resource-efficient foundation model programming} in which 
a task is automatically translated into an FMP that captures subtask dependencies and conditional control flow. \revise{The framework is modality-agnostic, making it directly applicable to any domain — text, vision, speech, or multimodal — as long as the task can be expressed as an FMP. } Each submodule of the program is then assigned to one of several backend models, differing in resource cost and capability. For example, consider the scenario in Figure \ref{fig:background}, in which a visual question answering (VQA) system receives the query ``\textit{Is there a cat sitting or laying on a laptop keyboard?}'' Here, our method generates a program that uses a small, inexpensive object detection model to check whether both a cat and a laptop are present. Only if that condition is met does it invoke a more powerful vision-language model (VLM) for finer-grained reasoning. 

We specifically focus on ``streaming'' tasks in which the system repeatedly solves a task --- for example, answering a question --- on a \emph{sequence} of inputs. An answer is generated without prior knowledge of the ground truth, then the system receives ground-truth feedback after each answer. In such settings, the cost of using a monolithic model is proportional to the number of inputs processed. By contrast, our approach uses the feedback from the early answers to learn a \emph{policy} that dynamically selects which backend model to invoke for each subtask, conditioned on the program input. Specifically, we use a combination of a structured REINFORCE estimator and Thompson Sampling to learn this policy. 

%and optional branching points. Next, 
%it uses a combination of a structured REINFORCE algorithm and gradient-based Thompson Sampling to 

% In such a setting, our approach first calls an LLM to generate a ``program sketch" that captures subtask dependencies, conditional control flow, and optional branching points. Next, it uses a combination of a structured REINFORCE algorithm with gradient-based Thompson Sampling to learn a \emph{policy} that dynamically selects which FM backend to invoke for each subtask, 
% In the specific example, this learned policy uses a small, inexpensive object detection model to check whether both a cat and a laptop are present. Only if that condition is met does it invoke a more powerful vision-language model (VLM) for finer-grained reasoning. 

While existing routing or cascading strategies \citep{chen2023frugalgpt, shnitzer2023large, lu2023routing, nie2024online} attempt to reduce large language model (LLM) inference overheads by switching between model sizes, they do not exploit the rich structural dependencies that arise in complex, compositional workflows. By contrast, our programs make these dependencies explicit, opening up opportunities for more flexible resource optimization.

Given the lack of standard benchmarks for resource-efficient sequential decision-making, we evaluate our approach on two newly introduced benchmarks: (1) a streaming binary VQA benchmark, where the questions require yes/no answers, spanning 33 compositional reasoning tasks with over 2,000 annotated images per task; and (2) a streaming open-form VQA benchmark, involving diverse questions with a broader answer space, covering 50 tasks with 500 annotated images per task. Experimental results show that our FMP-based system consistently reduces inference costs by 50\% to 98\% compared to one-size-fits-all baselines, without compromising task accuracy.

In summary, our contributions are as follows:
\begin{itemize}[leftmargin=0.15in,itemsep=0em]
    \item We propose the use of \emph{foundation model programs} as a flexible approach to 
    %exploiting task structure in 
    cost-efficient inference for complex, multimodal workflows.
    \item We give a specific method for learning such programs in a sequential decision-making setting. The highlight of the method is an online resource allocation method that systematically trades off the resource consumption and performance of models in an input-dependent way.  
    %explores the Pareto frontier between resource consumption and performance.
    \item We release two streaming benchmarks for binary and open-form VQA, reflecting real-world tasks where inputs arrive sequentially at scale and resource-efficiency is key.
    \item We show empirical results on these benchmarks, which demonstrate that our program-based approach can achieve up to 98\% cost savings with minimal accuracy degradation.%, indicating strong potential for practical deployment.
\end{itemize}

\section{Problem Formulation}
\label{sec:problem}
\paragraph{Foundation Model Programs} We consider programs written in a language such as Python, potentially synthesized by an LLM. The programs are \emph{neurosymbolic} because they interleave symbolic control flow with calls to a fixed set of \emph{generic neural functions} $F = \{f_{1}, f_{2}, \ldots, f_{K}\}$, where each $f_k$ denotes a high-level functionality (\eg object detection, visual question answering, natural language understanding). For example, the program in Figure~\ref{fig:background} makes calls to two generic neural functions \texttt{ImagePatch.find()} and \texttt{ImagePatch.vqa()}. In practice, we are interested in the case where the neural functions are implemented via foundation models. Hence, we use the term \emph{foundation model program} (FMP) to refer to such programs. 

Each generic function $f_k$ in an FMP has an associated set of $n_k$ \emph{backend models} that can be used to implement it, namely $M_{k} \;=\; \bigl\{m_{k,1},\;m_{k,2},\;\ldots,\;m_{k,n_{k}}\bigr\}$, where each backend $m_{k,j}$ has different trade-offs between accuracy and computational cost.  These backends may span a spectrum of models, from lightweight task-specific models to large, general-purpose language or multimodal models. Without loss of generality, we assume that the cost of invoking a backend is fixed and independent of the specific input it processes.

Now, assume that we analyze the program and produce an arbitrary ordering for the total of $N$ calls to generic neural functions, such that $k_i$ is the identity of the generic neural function associated with the $i$-th call in the program. Note that a program can call the same neural function multiple times with different inputs or arguments. The list of neural functions called by the program is then $\langle f_{k_{1}}, \;f_{k_{2}}, \;\ldots,\;f_{k_{N}} \rangle$.  In our example program, the list is $\langle \texttt{ImagePatch.find()}, \texttt{ImagePatch.find()}, 
\texttt{ImagePatch.vqa()} \rangle$.

Further, assume the host programming language has a runtime in which we are able to dynamically assign each of the $N$ calls to generic neural functions to a particular backend. Let $j_i$ denote which of the backends is selected for the $i$-th generic neural function call, where $j_i \in \{1, ..., n_{k_i}\}$.  That is, we select the $j$-th available neural backend $m_{k_i, j_i}$ for the $i$-th generic neural function call. 

Thus, we can customize the behavior of an FMP to optimize for accuracy and runtime cost on a specific program input, by choosing a specific list of backends $\vec{v} = \langle m_{k_1,j_1}, ..., m_{k_N,j_N}\rangle$.  We call $\vec{v}$ a \emph{program configuration vector}.  On input $x$, we use $p (x | \vec{v})$ to denote the output of the program, given that program configuration vector $\vec{v}$ was chosen.

\paragraph{Generality} This framework generalizes the formulation of existing LLM resource management methods, including routing and cascading.  Routing, as it is typically defined, is the problem of choosing a specific backend for a simple program that consists of a single generic neural function call: \texttt{lambda x :} $f(\texttt{x})$.  Cascading can be implemented as a foundation model program with nested \texttt{if}-\texttt{else} statements.

\paragraph{Task objective}
The idea of using FMPs for resource use optimization can be instantiated in a wide range of problems. In this paper, we focus on settings in which the goal is to solve a task --- for example, answer a question --- on a 
sequence of input-output pairs $\{x_t, y_t\}_{t=1}^T$. We assume that the structure of the programs we use only depends on the overall task and not the specific inputs. Therefore, on the input $x_t$ for time step $t$, we only need to decide on a suitable program configuration vector $\vec{v}_t$. 
%A foundation model program is executed repeatedly on a sequence of input-output pairs $\{x_t, y_t\}_{t=1}^T$. 
%At each time step $t$, the goal is to select a program configuration vector $\vec{v}_t$ that is well-suited to the input $x_t$, 
We want $\vec{v}_t$ to be 
such that the program execution output $p(x_t \mid \vec{v}_t)$ approximates the ground truth $y_t$, while minimizing execution cost. To capture this trade-off, we define the following reward function:
\begin{equation*}
R\bigl(\vec{v}_{t},\, x_{t},\, y_{t}\bigr) 
\;=\; 
-\,\mathcal{L}\bigl(p(x_t | \vec{v}_t),\, y_{t}\bigr)
\;-\;\lambda\,\mathcal{C}\bigl(\vec{v}_{t}\bigr),
\end{equation*}

where $\mathcal{L}$ quantifies the output discrepancy between $p(x_t \mid \vec{v}_t)$ and the ground truth $y_t$, $\mathcal{C}$ represents the actual computational cost incurred when running the program with configuration $\vec{v}_t$ on input $x_t$,  and $\lambda>0$ is a trade-off weighting factor. Importantly, due to control flow in the program, not all neural backends specified in $\vec{v}_t$ may be invoked on a given input; $\mathcal{C}$ accounts only for the cost of the operations actually executed.

Over $T$ time steps, the objective is to learn a \emph{policy} $\pi$ that maps each input $x_{t}$ to a program configuration vector $\vec{v}_{t}$. Let $\Pi$ be the space of such policies. We seek to solve the problem 
\begin{equation}\label{eq:problem}
\max_{\pi \in \Pi} \;\sum_{t=1}^T 
R\bigl(\vec{v}_{t},\, x_{t},\, y_{t}\bigr)
\quad 
\text{subject to} 
\quad 
\vec{v}_{t} \;=\;\pi\!\bigl(x_{t}\bigr).
\end{equation}
Importantly, we require this optimization problem to be solved \emph{online}. That is, we 
% \paragraph{Online learning} Further, we consider an online setting in which i
assume that our inputs arrive sequentially and require decisions to be made without prior knowledge of the ground truth. Only after the selected configuration is executed and the output $p(x_t \mid \vec{v}_t)$ is produced is the ground truth $y_t$ revealed and the reward $R(\vec{v}_t, x_t, y_t)$ computed. 
\section{Methodology}

Our approach to addressing this problem consists of two key phases: \emph{offline code generation} and \emph{online resource allocation}.

\subsection{Offline Code Generation }
We begin by synthesizing a foundation model program $p$ from the user specification using an LLM. This process yields a task-specific program sketch, including a sequence of generic neural function calls $\{f_{k_1}, f_{k_2}, \ldots, f_{k_N}\}$, which defines the high-level structure of the computation, while the backend selection for these neural functions is determined online. 

\revise{The correctness of the FMP is critical, as it defines the entire computational workflow for the task. To ensure transparency and verifiability, FMPs are constructed with an explicit neurosymbolic representation, making them straightforward to inspect, validate, and debug. Although we employ an LLM to generate these programs, our framework is designed to be agnostic to the source of the FMP and fully supports human-authored workflows.}

\subsection{Online Resource Allocation}
After offline synthesis, the main challenge is to select a program configuration vector $\vec{v}_t$ for each input $x_t$, dynamically assigning a neural backend $m_{k_i,j_i} \in M_{k_i}$ to each function call $f_{k_i}$. The space of possible configurations grows combinatorially with $N$, making exhaustive search intractable. To address this, we propose a structured policy that decomposes this decision process into $N$ manageable sub-policies, one per function call. 

Specifically, for each function call $f_{k_i}$, we define a \emph{sub-reward} function:
\begin{equation*}
r_{k_i,j_i} = -\lambda \mathcal{C}(m_{k_i,j_i}) - \frac{1}{N}\mathcal{L}(p(x_t|\vec{v}_t), y_t)\quad \text{where }m_{k_i,j_i} \in \vec{v_t}.
\end{equation*}

This sub-reward decomposes the global reward $R(\vec{v}_t, x_t, y_t)$ into structured contributions. It integrates the local computational cost $\mathcal{C}(m_{k_i,j_i})$ associated with backend $m_{k_i,j_i}$, and a portion of the predictive loss $\mathcal{L}(p(x_t|\vec{v}_t), y_t)$, that is determined when the program $p$ is executed with the entire configuration $\vec{v}_t$. 

To model these rewards, we define a \emph{subpolicy} $\pi_{k_i}$ with learnable parameters~$\theta_{k_i}$.  Given the input $x_t$, subpolicy $\pi_{k_i}$ outputs a reward prediction
\begin{equation*}
r'_{k_i,j_i} = \pi_{k_i}(m_{k_i,j_i}|x_t;\theta_{k_i})\quad \text{for each }m_{k_i,j_i} \in M_{k_i}. 
\end{equation*}

This structured design simplifies the decision space: rather than jointly optimizing over all $n_{k_1} \times \cdots \times n_{k_N}$ backend combinations, we train $N$ separate subpolicies.  Each subpolicy is specialized to one of the $N$ function calls, thereby simplifying the optimization process, enabling parallel learning, and reducing unwanted interference across calls. \revise{This per-call decomposition is critical for a precise credit assignment. In a monolithic joint policy, it is difficult to attribute rewards correctly, especially in the presence of control flow where some function calls may be conditionally skipped. Our structured approach resolves this by updating a subpolicy only when its corresponding function is executed, ensuring that reward signals are correctly aligned with the decisions that caused them.}

\paragraph{Gradient-based Thompson Sampling} To balance exploration and exploitation in the online setting, decisions are made using Thompson Sampling~\cite{zhang2020neural} instead of greedily selecting the FM backend with the highest predicted reward. For each backend $m_{k_i,j_i} \in M_{k_i}$, the subpolicy samples a reward from a normal distribution:
\begin{equation*}
\hat{r}_{k_i,j_i} \sim \mathcal{N}\bigl(r'_{k_i,j_i},\, (\nu \cdot \sigma_{k_i,j_i})^2\bigr),
\end{equation*}
where $\sigma_{k_i,j_i} = \sqrt{\sum_l \frac{g_{k_i,j_i,l}^2}{\mathbf{U}_{k_i,l}}}$ quantifies uncertainty in the reward prediction. Here, $l$ indexes each individual parameter of the subpolicy, $g_{k_i,j_i,l}$ is the gradient of the reward prediction with respect to parameter $l$, $\mathbf{U}_{k_i}$ tracks the accumulated gradient‐based parameter uncertainties, and $\nu$ scales the exploration. We select the backend with the highest sampled reward:
\begin{equation*}
j_i^* = \argmax_{j_i \in \{1,...,n_{k_i}\}} \hat{r}_{k_i,j_i}.
\end{equation*}
After selection, the uncertainty parameter $\mathbf{U}_{k_i}$ is updated to refine future exploration:
\begin{equation*}
\mathbf{U}_{k_i,l} \leftarrow \mathbf{U}_{k_i,l} + g_{k_i, j_i^*, l}^2,
\end{equation*}
where $g_{k_i, j_i^*, l}$ is the gradient of the selected backend $m_{k_i, j_i^*}$ with respect to parameter $l$. 

We repeat the above process for each $i=1,\dots,N$, allowing each subpolicy to choose one backend per function call, thereby yielding the program configuration vector at time step $t$: 
\begin{equation*}
\vec{v}_t = \bigl(m_{k_1, j_1^*}, m_{k_2, j_2^*}, \ldots, m_{k_N, j_N^*}\bigr).
\end{equation*}

\begin{algorithm}[t]
\caption{Structured REINFORCE Framework}
\label{algo:structured_reinforce}
\begin{algorithmic}
\STATE \textbf{Initialize:} Policy parameters $\theta_{k_i}$, uncertainty estimates $\mathbf{U}_{k_i}$, learning rate $\eta$, exploration factor $\nu$
\FOR {each input $x_t$}
    \FOR {each function call $f_{k_i}$ in program $p$}
        \STATE Predict reward $r'_{k_i,j_i}=\pi_{k_i}(m_{k_i,j_i}|x_t;\theta_{k_i})$ for each FM backend $m_{k_i,j_i}$
        \STATE Compute uncertainty $\sigma_{k_i,j_i} = \sqrt{\sum_l \frac{g_{k_i,j_i,l}^2}{\mathbf{U}_{k_i,l}}}$
        \STATE Sample adjusted reward for exploration: \quad $\hat{r}_{k_i,j_i} \sim \mathcal{N}\bigl(r'_{k_i,j_i},\, (\nu \cdot \sigma_{k_i,j_i})^2\bigr)$
        \STATE Select FM backend $m_{k_i, j_i^*}$ with highest sampled reward $\hat{r}_{k_i,j_i}$
        \STATE Update parameter uncertainties $\mathbf{U}_{k_i,l}$
    \ENDFOR
    \STATE Execute program $p$ with selected configuration $\vec{v}_t=(m_{k_1,j_1^*}, ..., m_{k_N,j_N^*})$
    \STATE Observe final reward $R(\vec{v}_t, x_t, y_t)$
    \FOR {each function call $f_{k_i}$}
        \STATE Compute policy gradient $\nabla_{\theta_{k_i}} \mathcal{J}(\pi_{k_i})$ based on observed reward
        \STATE Update policy parameters: \quad $\theta_{k_i} \leftarrow \theta_{k_i} - \eta \,\nabla_{\theta_{k_i}}\mathcal{J}(\pi_{k_i})$
    \ENDFOR
\ENDFOR
\end{algorithmic}
\end{algorithm}

\paragraph{Structured REINFORCE Algorithm}
We now describe the online learning of the subpolicies using only the global reward $R(\vec{v}_t, x_t, y_t)$ observed after execution. The learning objective for the overall policy $\pi = \{\pi_{k_1}, \dots, \pi_{k_N}\}$ is to maximize cumulative reward over $T$ episodes:
\begin{equation*}
\mathcal{J}(\pi) = \sum_{t=1}^T R\bigl(\vec{v}_t, x_t,y_t\bigr), \quad\text{where }\vec{v}_t = \bigl(m_{k_1, j_1^*}, m_{k_2, j_2^*}, \ldots, m_{k_N, j_N^*}\bigr).
\end{equation*}

Since the reward $R(\vec{v}_t, x_t, y_t)$ is equivalent to the aggregation of all sub-rewards:
\begin{align*}
    R(\vec{v}_t, x_t, y_t) &= -\,\mathcal{L}\bigl(p(x_t | \vec{v}_t),\, y_{t}\bigr)
\;-\;\lambda\,\mathcal{C}\bigl(\vec{v}_{t}\bigr)\\ 
 &=\sum_{i=1}^{N} \Bigr[ - \frac{1}{N}\mathcal{L}(p(x_t|\vec{v}_t), y_t) -\lambda \mathcal{C}(m_{k_i,j_i})  \Bigl] =\sum_{i=1}^{N}r_{k_i, j_i^*},
\end{align*}
we convert the learning objective into optimizing each subpolicy $\pi_{k_i}$ independently:
\begin{equation*}
\mathcal{J}\bigl(\pi_{k_i}\bigr) 
\;=\; 
\sum_{t=1}^T 
\mathbb{E}_{
\,j_i^*\,\sim\,\pi_{k_i}(\cdot\mid x_t)}
\Bigr[r_{k_i,j^*_i}\Bigl].
\end{equation*}

Because the program execution is non-differentiable due to control flow structures like conditional branches and loops \cite{kreikemeyer2023smoothing}, we employ the REINFORCE algorithm \cite{williams1992simple} to estimate policy gradients:
\begin{equation*}
\nabla_{\theta_{k_i}}
\mathcal{J}\bigl(\pi_{k_i}\bigr)
\;=\;
\sum_{t=1}^T
\mathbb{E}_{
\,j_i^*\,\sim\,\pi_{k_i}(\cdot\mid x_t)}\!
\Bigl[
\nabla_{\theta_{k_i}}\,
\log\,\pi_{k_i}\bigl(m_{k_i,j_i^*}\mid x_t;\,\theta_{k_i}\bigr)
\;\cdot\;
r_{k_i,j_i^*}
\Bigr].
\end{equation*}
Note that each $r_{k_i, j_i^*}$ depends on the full program execution but reflects a partial credit assignment for subpolicy $\pi_{k_i}$.

In practice, we approximate the expectation using $S$ sampled trajectories:
\begin{align*}
\nabla_{\theta_{k_i}}
\mathcal{J}\bigl(\pi_{k_i}\bigr)
\;\approx\;
\sum_{t=1}^T
\sum_{s=1}^S
\nabla_{\theta_{k_i}}\,
\log\,\pi_{k_i}
\bigl(m_{k_i,\,j_i^*}^{(s)}\mid x_t^{(s)};\,\theta_{k_i}\bigr)
\;\cdot\;
r_{k_i,\,j_i^*}^{(s)}.
\end{align*}

The sub-policies are periodically trained to stabilize learning:
\begin{equation*}
\theta_{k_i} \leftarrow \theta_{k_i} - \eta \nabla_{\theta_{k_i}} \mathcal{J}(\pi_{k_i}),
\end{equation*}
where $\eta$ is the learning rate. 

The overall framework is detailed in Algorithm \ref{algo:structured_reinforce} with a walk-through example in Appendix Figure \ref{fig:demo}. This methodology provides a tractable solution to the online resource allocation challenge, with a theoretical no-regret guarantee provided in Appendix \ref{sec:app_proof}. By separating offline synthesis from online optimization and employing a decomposed policy structure, we achieve both flexibility and efficiency in program execution.
% Given an input, the structured policy predicts rewards for different backend choices at each function call, incorporating a gradient-based Thompson Sampling to balance exploration and exploitation. This structured framework ensures modular optimization, enabling independent learning of function-specific policies while maintaining tractability. We also provide a no-regret theoretical guarantee for the algorithm in Appendix \ref{sec:app_proof}. 
\section{Benchmark}
Motivated by the need for structured, sequential evaluation beyond the single-image-per-query setups typical of existing visual question answering (VQA) datasets \cite{goyal2017making}, we introduce two novel streaming VQA benchmarks. \revise{These benchmarks are designed to reflect real-world latency-critical deployments by simulating input streams, long-tail difficulty, class imbalance, and adversarial distractors. While partially synthetic, they are constructed using real data (e.g., COCO, GQA) and validated through LLM and human annotations to ensure complexity and robustness.}

Our first \textbf{Streaming Binary VQA} benchmark focuses on yes/no question answering, a task commonly studied in previous works \cite{antol2015vqa, zhang2016yin, hudson2019gqa}. In this benchmark, systems are challenged to determine whether a sequence of images satisfies complex, compositional queries. These queries incorporate diverse reasoning types—spatial (\eg “Is there a person riding a bicycle next to a bus on the street?”), logical (\eg “Are there people riding bikes, scooters, or motorcycles while holding or using umbrellas?”), and numerical (\eg “Are there at least four horses on a beach?”)—to better reflect real-world reasoning demands. The final benchmark includes 33 queries with more than 2000 annotated images for each query, featuring a realistic class imbalance setup. Further details on the benchmark construction and evaluation are provided in Appendix~\ref{sec:appendix_benchmarks_a}.

\begin{figure*}[t]
     \centering
     \includegraphics[width=\textwidth]{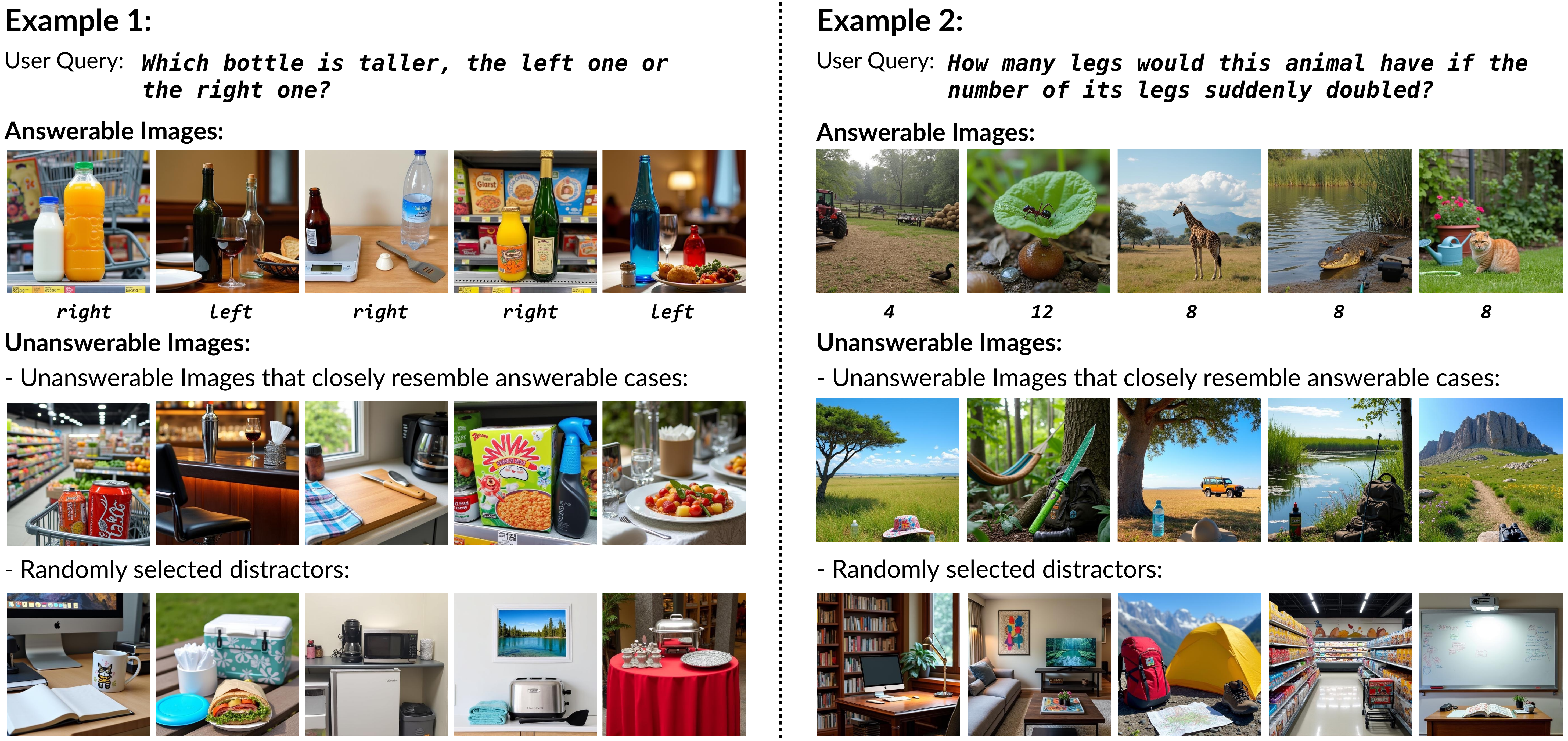}
     \caption{Examples of the synthetic images in the Streaming Open-form VQA benchmark. } 
     \label{fig:example}
\end{figure*}

Our second benchmark, \textbf{Streaming Open-form VQA}, evaluates a system’s ability to answer open-form questions for a sequence of input images. This benchmark spans five reasoning categories: spatial (\eg “What is in the jar to the left of the juice?”), logical (\eg “What is the black object on the desk that is not electronic?”), numerical (\eg “How many extra bottles of beer do we need to make it a half dozen?”), comparative (\eg “Which bottle is taller, the left one or the right one?”), and external knowledge reasoning (\eg “How many states are there in the country whose flag is shown?”). Images are generated using a diffusion model with a dedicated pipeline to ensure diversity and quality control. To evaluate model robustness, we also introduce unanswerable images that are visually similar to the query but semantically invalid for answering. The final benchmark includes 50 queries with 500 annotated images per query. Complete details of the image generation pipeline, neurosymbolic program synthesis, and evaluation metrics (exact match accuracy) are described in Appendix~\ref{sec:appendix_benchmarks_b}.

\section{Experiments}
\subsection{Implementation Details and Experimental Setups}
We implement the structured policy using ResNet-18 with $\sim$11M parameters \cite{he2016deep}, ensuring the policy training and inference overheads do not exceed the resource savings. During evaluation, our experiments use varying sizes of GLIP \cite{li2022grounded} and GroundingDINO \cite{liu2023grounding} for object detection, OFA \cite{wang2022ofa}, BLIP-2 \cite{li2023blip}, and Qwen2.5-VL series \cite{bai2025qwen2} for visual-language understanding, and Llama 3 series \cite{dubey2024llama} as the LLMs. We implement an instrumentation system that analyzes the programs to determine function calls, modifies the abstract syntax tree (AST) to inject program configurations, and executes the modified program while collecting performance metrics including the execution traces and the computational costs. 

\subsection{Baselines}
We compare our system against
several baselines to establish its effectiveness:

\paragraph{Single MLLMs} We evaluate our system against the state-of-the-art multimodal LLMs (MLLMs) that integrate both vision and language reasoning capabilities \cite{bai2025qwen2}. 

\paragraph{MLLM Routing} As an alternative adaptive strategy, a multi-armed bandit dynamically routes user queries to multimodal LLMs of varying sizes with different cost-accuracy trade-offs based on estimated rewards, balancing exploration and exploitation \cite{nguyen2024metallm, li2025llm}. However, it does not account for the task structures in user queries.

\paragraph{Static FM Program Configurations} A common approach to resource management is to use a fixed, pre-determined configuration of foundation models for the FM programs without dynamic backend selection. Given the combinatorial space of configurations, we implement two variants: one using the cheapest FM configurations and another using the most expensive configurations.

\paragraph{Pareto-Random Routing}
Following the standard practice in prior works \cite{hu2024routerbench, jitkrittum2025universal}, we employ a straightforward yet effective Pareto-random routing strategy through linear interpolation. We implement this approach separately for two scenarios: multimodal LLMs and static FM program configurations. 
% Specifically, given a parameter $t \in [0, 1]$, our router probabilistically selects either the cheaper or more expensive model (or configuration), choosing the former with probability $t$ and the latter with probability $1 - t$. 

In the experiments reported in the main paper, we consistently use Qwen2.5-VL (3B and 72B) for both the Single MLLM and LLM Routing baselines. The FM program backends consist of Grounding-DINO Tiny (172M) and Base (224M) for object detection, along with Qwen2.5-VL 3B and 72B for vision-language understanding.

\subsection{Results}
\begin{figure*}
    \centering
    \begin{subfigure}[t]{.3\textwidth}
        \centering
        \includegraphics[height=1.7in]{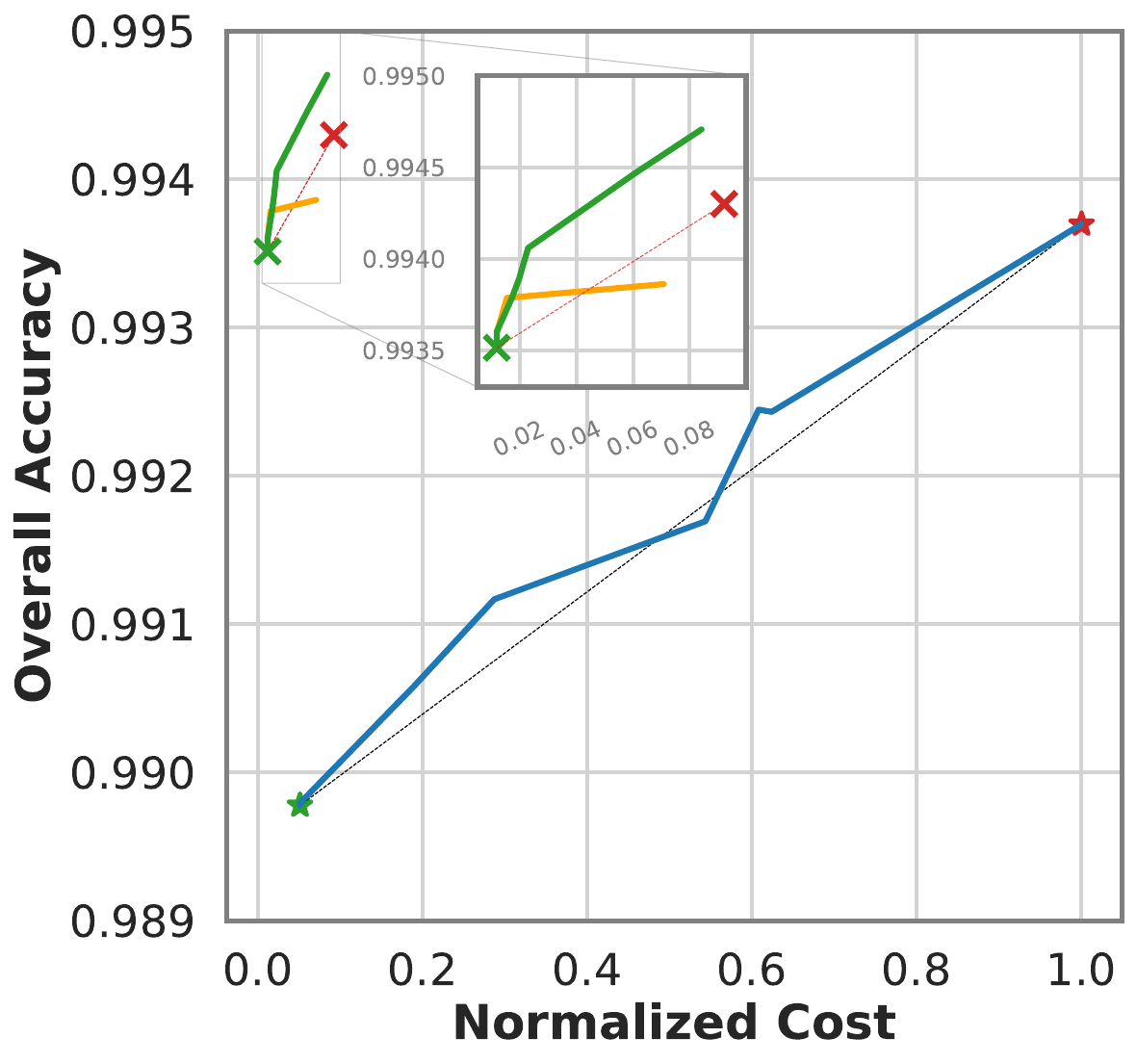}
    \end{subfigure}%
    ~ 
    \begin{subfigure}[t]{.73\textwidth}
        \centering
        \includegraphics[height=1.7in]{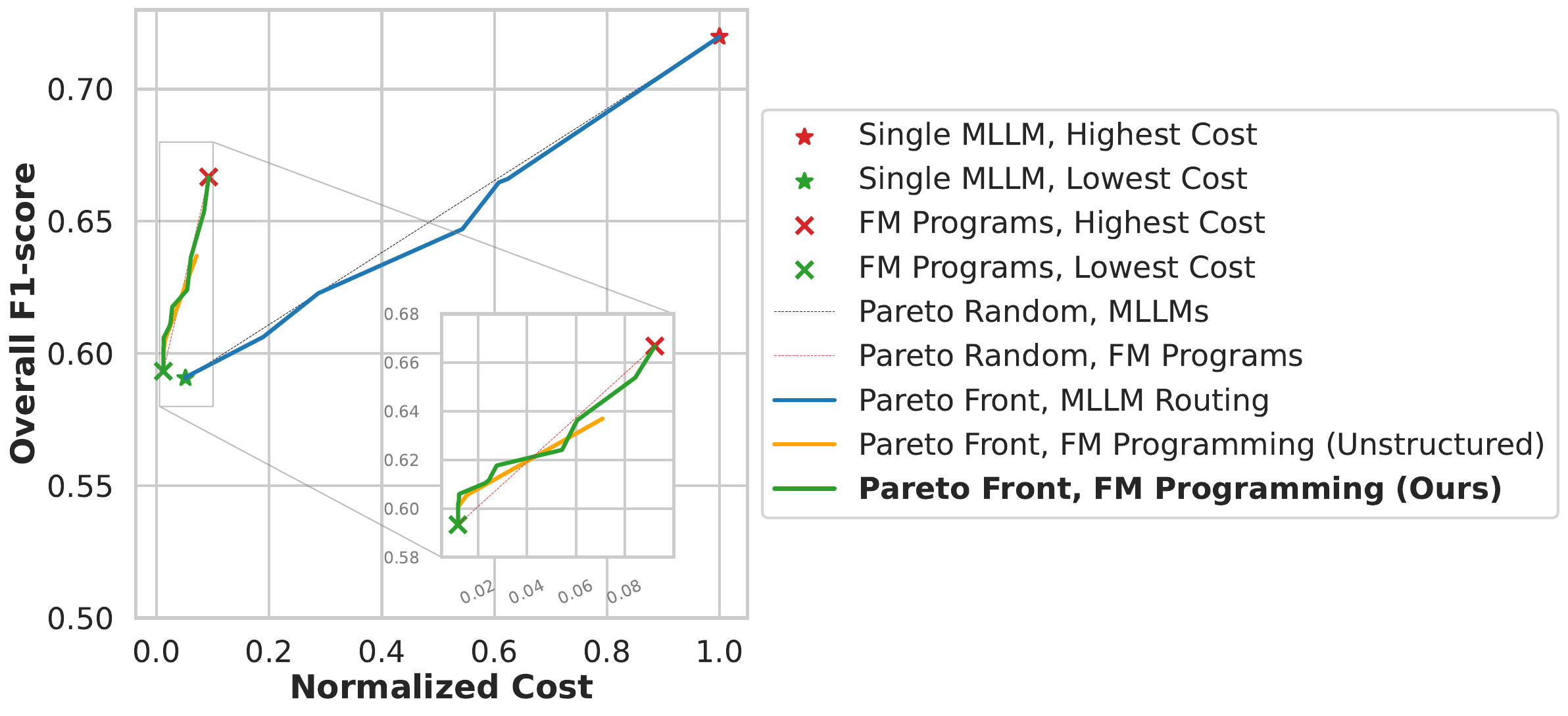}
    \end{subfigure}

\caption{Experimental results on the Streaming Binary VQA benchmark. Costs are normalized based on the inference costs of the most expensive MLLM, \ie Qwen2.5-VL 72B.}
\label{fig:verification_results}
\end{figure*}

% \begin{figure*}
% \centering
% \includegraphics[width=\textwidth]{figures/combined.pdf}
% \caption{Experimental results on the (a) \textit{Streaming Binary VQA} benchmark, and (b) \textit{Streaming Open-form VQA} benchmark. Costs are normalized based on the inference costs of the most expensive MLLM, \ie Qwen2.5-VL 72B.}
% \label{fig:verification_results}
% \end{figure*}

\paragraph{Streaming Binary VQA} 
As shown in Figure~\ref{fig:verification_results}, the FM Programming approach achieves accuracy comparable to the largest MLLM, while reducing computational costs by over 98\%. This striking efficiency gain stems from its ability to exploit task structure: FM Programs use lightweight object detection modules to filter out most negative samples, drastically reducing expensive MLLM inference.

Beyond cost savings, FM Programming consistently delivers superior cost-accuracy and cost-F1 trade-offs, particularly in the low-cost regime. The Pareto frontier of FM Programming (green line) significantly outperforms both the Pareto Random baseline for FM Programs (red dashed line) and the LLM routing baseline (blue line), illustrating the effectiveness of structured decision-making via the Structured REINFORCE algorithm.

% As shown in Figure \ref{fig:verification_results}, the FM Programming approach achieves accuracy comparable to the largest MLLM, while reducing computational costs by over 98\%. By leveraging task structure, FM programs filter out the majority of negative samples using lightweight object detection models, significantly reducing the need for expensive MLLM inference. Additionally, the Pareto Front of FM Programming (green line) consistently outperforms the Pareto Random baseline (red dashed line), highlighting the effectiveness of the Structured REINFORCE algorithm in exploring the cost-performance Pareto Frontier.

\begin{figure*}[t]
    \centering
    \includegraphics[height=1.9in]{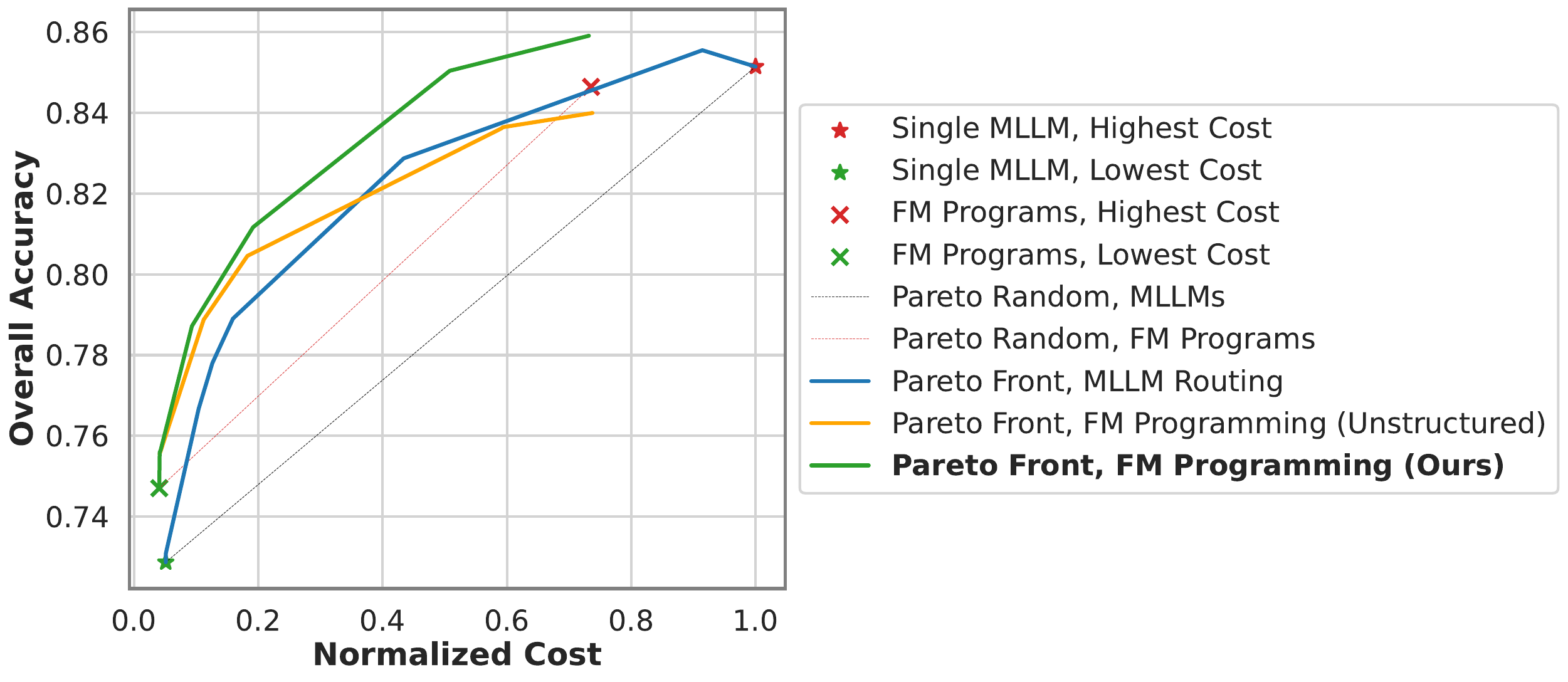}
\caption{Experimental results on the Streaming Open-form VQA benchmark. Costs are normalized based on the inference costs of the most expensive MLLM, \ie Qwen2.5-VL 72B.}
\label{fig:vqa_results}
\end{figure*}

Compared to the LLM routing baseline, FM Programming achieves a 6\% improvement in F1 score at similar computational cost (10\% of the largest MLLM). While its maximum F1 score is lower than that of the Single MLLM and LLM routing baselines, this is due to conservative thresholds in the object detection modules that prioritize precision over recall to minimize costly false positives. As shown in Appendix Figure~\ref{fig:verification_recall_precision}, the learned policy increasingly favors precision over recall as the cost budget grows. In practice, this trade-off can be adjusted by tuning the reward function used in Structured REINFORCE.

% Compared to the LLM routing baseline, our method yields a 4\% improvement in accuracy and a 6\% increase in F1 score at equivalent computational costs. Although the maximum F1 score is slightly lower than the Single MLLM and LLM routing baselines, this is due to conservative thresholds in the object detection models, which prioritize precision over recall to minimize costly false positives. As shown in Figure \ref{fig:verification_recall_precision}, as the cost budget increases, the learned policy tends to further favor reducing false positives (higher precision), sometimes at the expense of recall. This trade-off can be adjusted by modifying the reward function used in Structured REINFORCE.

To evaluate robustness, we conduct additional experiments using an alternative set of FM backends. As shown in Appendix Figure~\ref{fig:verification_add}, despite reduced backend capabilities leading to lower raw accuracy and F1 scores, FM Programming still maintains clear Pareto dominance over the baseline. This demonstrates its strong generalizability and adaptability across different FM backend configurations.

% To further validate the robustness of our approach, we conduct additional experiments with an alternative set of FM backends. As shown in Appendix Figure \ref{fig:verification_add}, despite the reduced accuracy and F1 scores due to limitations in backend capabilities, the FM Programming method consistently achieves superior cost-accuracy and cost-F1 trade-offs compared to the Pareto Random baseline.

\paragraph{Streaming Open-form VQA} We further evaluate our method on the more challenging Streaming Open-form VQA benchmark. As shown in Figure~\ref{fig:vqa_results}, FM Programs (red dashed line) consistently outperform MLLMs (black dashed line) by over 2\% in accuracy at equivalent cost, showcasing the advantage of modular programs over end-to-end models in leveraging task structures.

% \paragraph{Streaming Open-form VQA Results} As shown in Figure \ref{fig:vqa_results}, our proposed FM Programming approach demonstrates strong performance on the more challenging open-form VQA benchmark. When comparing the Pareto frontiers, FM programs (red dashed line) outperform MLLMs (black dashed line) by more than 2\% in accuracy at equivalent costs. This highlights the advantage of modular programs over end-to-end models in leveraging task structures.

Building on this, the FM Programming approach (purple line) effectively expands the Pareto frontier. Through dynamic backend allocation with Structured REINFORCE, it achieves up to 50\% cost savings without sacrificing performance compared to the largest MLLM, and in some cases even outperforms the MLLM while reducing cost by 30\%.

Notably, FM Programming offers high marginal returns in low-cost regimes — small increases in cost yield substantial accuracy gains.  It clearly surpasses the Pareto Random baseline and consistently outperforms the LLM Routing method, providing better cost-performance trade-offs. This demonstrates that the proposed FM Programming framework is effective in fully leveraging the task and input structure for optimal cost-efficiency.

\paragraph{Ablation Study} \revise{To assess our structured policy design, we compare it against an unstructured policy variant that jointly optimizes all backend selections. As shown in Figures \ref{fig:verification_results} and \ref{fig:vqa_results}, the unstructured policy (yellow line) significantly underperforms on both benchmarks, due to its inability to manage conditional branches and accurately assign rewards to local decisions, particularly when the resource budgets increase. This highlights the critical role of our Structured REINFORCE algorithm in enabling precise credit assignment and effective resource allocation.}

These results emphasize that FM Programming is not only cost-efficient but also highly adaptive and scalable. Its ability to generalize across both binary and open-form VQA tasks suggests strong potential for real-world deployment in latency- and resource-constrained scenarios.

% FM Programming (purple line) further improves on this by dynamically allocating backends with the structured REINFORCE algorithm. It can navigate the Pareto frontier of FM programs more efficiently. As a result, it saves about 50\% in cost without any loss in performance compared to the largest evaluated MLLM. In some cases, it even outperforms the MLLM while reducing cost by around 30\%.
\section{Related Works}

\subsection{Multi-modal Reasoning}
Multi-modal reasoning tasks, such as visual question answering, require integrating vision and language to address complex queries. End-to-end models, powered by large pre-trained transformers \cite{deitke2024molmo, bai2025qwen2}, achieve high accuracy but often come with significant computational costs and limited interpretability. In contrast, modular approaches like ViperGPT \cite{suris2023vipergpt} and VisProg \cite{gupta2023visual} decompose complex tasks into smaller, executable programs, enhancing flexibility and interpretability by dynamically composing models based on task needs. 
Compared to the existing methods that rely on fixed programs, our work extends this modular paradigm with a particular focus on cost-efficiency, introducing dynamic FM backend selection to handle varying input complexities for achieving the Pareto frontier between resource and performance. 

\subsection{Foundation Model Inference Cost Optimization}
Optimizing foundation models, such as LLMs and MLLMs, is essential for deployment in resource-constrained environments. \revise{Common techniques include distillation \cite{hinton2015distilling, gu2023knowledge} and quantization \cite{gholami2022survey, lin2024awq} that operate at the model level, and speculative decoding \cite{leviathan2023fast, miao2023specinfer} and mixture-of-experts routing \cite{fedus2022switch, huang2024harder} that operate at the token level. In contrast, our approach works at the task level, allowing dynamic resource allocation across multiple subtask backends, making it complementary to these strategies. }

More recently, Compound AI systems have emerged, leveraging multiple model backends to optimize inference costs. Among these, LLM routing selects the best model for a query to balance cost and quality \cite{lu2023routing, hu2024routerbench}; Cascading uses a series of models, starting with simpler ones for easy tasks and escalating to complex ones as needed \cite{chen2023frugalgpt, nie2024online}. 
Although effective, these methods generally lack integration with task-specific reasoning structures. In contrast, our approach combines task decomposition with dynamic backend selection in FM programs, enabling resource allocation that adapts to both input complexity and task structures for cost saving. 

\section{Conclusion}
We introduced the first framework to use foundation model programs in optimizing the trade-off between task performance and resource consumption. By decomposing complex, multi-modal reasoning tasks into programs with control flow and dynamically selecting FM backends in online setups, our method can navigate the Pareto frontier of trade-offs between resource use and performance in real-time, in a way that exploits both task and input structure. As demonstrated in our experiments, this strategy can deliver substantial computational savings with minimal performance degradation. 

Many directions of future work remain open. The idea of using FMPs for resource-efficient inference goes beyond the tasks we considered; in particular, we believe they can be deployed in broader agentic and multi-agent applications. Another direction is to explore more advanced program synthesis techniques, such as ones that jointly learn program structure and configurations.

%coordination strategies for multi-agent FM programming. 

\section*{Limitations}
\revise{Our framework's performance relies on several assumptions: (a) The initial program structure must be correct -- an illogical workflow will lead to sub-optimal performance, regardless of backend allocation; (b) The policy is constrained by the capabilities of the available FM backends. If the backend pool is too weak for a given subtask, even optimal routing may not achieve the desired accuracy, as shown in Appendix Figure \ref{fig:verification_add}; (c) The online learning algorithm requires immediate ground-truth feedback, which may not be available in real-world settings; (d) We also do not explicitly model error propagation, where an early mistake from a lightweight module can impact downstream steps.}

\section*{Acknowledgement}
\revise{We thank the anonymous reviewers for their valuable comments and suggestions in enhancing the paper. This research was supported by the NSF under grant numbers CCF-1918651, 2008240, 2131294, and 2212557, NSF PPoSS award \#2316161, ARO award \#W911NF-21-1-0009, DARPA award \#HR00112320018, NIH CTSA award \#UM1TR004906, and US DOT Tier-1 UTC CYBER-CARE grant \#69A3552348332.}
\clearpage
\bibliography{ref}

\begin{thebibliography}{55}
\providecommand{\natexlab}[1]{#1}
\providecommand{\url}[1]{\texttt{#1}}
\expandafter\ifx\csname urlstyle\endcsname\relax
  \providecommand{\doi}[1]{doi: #1}\else
  \providecommand{\doi}{doi: \begingroup \urlstyle{rm}\Url}\fi

\bibitem[Achiam et~al.(2023)Achiam, Adler, Agarwal, Ahmad, Akkaya, Aleman, Almeida, Altenschmidt, Altman, Anadkat, et~al.]{achiam2023gpt}
Josh Achiam, Steven Adler, Sandhini Agarwal, Lama Ahmad, Ilge Akkaya, Florencia~Leoni Aleman, Diogo Almeida, Janko Altenschmidt, Sam Altman, Shyamal Anadkat, et~al.
\newblock Gpt-4 technical report.
\newblock \emph{arXiv preprint arXiv:2303.08774}, 2023.

\bibitem[Agarwal et~al.(2021)Agarwal, Kakade, Lee, and Mahajan]{agarwal2021theory}
Alekh Agarwal, Sham~M Kakade, Jason~D Lee, and Gaurav Mahajan.
\newblock On the theory of policy gradient methods: Optimality, approximation, and distribution shift.
\newblock \emph{Journal of Machine Learning Research}, 22\penalty0 (98):\penalty0 1--76, 2021.

\bibitem[AgentLego(2023)]{contributors7agentlego}
Contributors AgentLego.
\newblock Agentlego: Open-source tool api library to extend and enhance llm agents, december 2023.
\newblock \emph{URL https://github. com/InternLM/agentlego}, 2023.

\bibitem[Antol et~al.(2015)Antol, Agrawal, Lu, Mitchell, Batra, Zitnick, and Parikh]{antol2015vqa}
Stanislaw Antol, Aishwarya Agrawal, Jiasen Lu, Margaret Mitchell, Dhruv Batra, C~Lawrence Zitnick, and Devi Parikh.
\newblock Vqa: Visual question answering.
\newblock In \emph{Proceedings of the IEEE international conference on computer vision}, pp.\  2425--2433, 2015.

\bibitem[Bai et~al.(2025)Bai, Chen, Liu, Wang, Ge, Song, Dang, Wang, Wang, Tang, et~al.]{bai2025qwen2}
Shuai Bai, Keqin Chen, Xuejing Liu, Jialin Wang, Wenbin Ge, Sibo Song, Kai Dang, Peng Wang, Shijie Wang, Jun Tang, et~al.
\newblock Qwen2. 5-vl technical report.
\newblock \emph{arXiv preprint arXiv:2502.13923}, 2025.

\bibitem[Ban et~al.(2021)Ban, He, and Cook]{ban2021multi}
Yikun Ban, Jingrui He, and Curtiss~B Cook.
\newblock Multi-facet contextual bandits: A neural network perspective.
\newblock In \emph{Proceedings of the 27th ACM SIGKDD Conference on Knowledge Discovery \& Data Mining}, pp.\  35--45, 2021.

\bibitem[{Black Forest Labs}(2023)]{flux2023}
{Black Forest Labs}.
\newblock Flux.
\newblock \url{https://github.com/black-forest-labs/flux}, 2023.

\bibitem[Chen et~al.(2019)Chen, Wang, Pang, Cao, Xiong, Li, Sun, Feng, Liu, Xu, et~al.]{chen2019mmdetection}
Kai Chen, Jiaqi Wang, Jiangmiao Pang, Yuhang Cao, Yu~Xiong, Xiaoxiao Li, Shuyang Sun, Wansen Feng, Ziwei Liu, Jiarui Xu, et~al.
\newblock Mmdetection: Open mmlab detection toolbox and benchmark.
\newblock \emph{arXiv preprint arXiv:1906.07155}, 2019.

\bibitem[Chen et~al.(2023)Chen, Zaharia, and Zou]{chen2023frugalgpt}
Lingjiao Chen, Matei Zaharia, and James Zou.
\newblock Frugalgpt: How to use large language models while reducing cost and improving performance.
\newblock \emph{arXiv preprint arXiv:2305.05176}, 2023.

\bibitem[Deitke et~al.(2024)Deitke, Clark, Lee, Tripathi, Yang, Park, Salehi, Muennighoff, Lo, Soldaini, et~al.]{deitke2024molmo}
Matt Deitke, Christopher Clark, Sangho Lee, Rohun Tripathi, Yue Yang, Jae~Sung Park, Mohammadreza Salehi, Niklas Muennighoff, Kyle Lo, Luca Soldaini, et~al.
\newblock Molmo and pixmo: Open weights and open data for state-of-the-art multimodal models.
\newblock \emph{arXiv preprint arXiv:2409.17146}, 2024.

\bibitem[Dubey et~al.(2024)Dubey, Jauhri, Pandey, Kadian, Al-Dahle, Letman, Mathur, Schelten, Yang, Fan, et~al.]{dubey2024llama}
Abhimanyu Dubey, Abhinav Jauhri, Abhinav Pandey, Abhishek Kadian, Ahmad Al-Dahle, Aiesha Letman, Akhil Mathur, Alan Schelten, Amy Yang, Angela Fan, et~al.
\newblock The llama 3 herd of models.
\newblock \emph{arXiv preprint arXiv:2407.21783}, 2024.

\bibitem[Fedus et~al.(2022)Fedus, Zoph, and Shazeer]{fedus2022switch}
William Fedus, Barret Zoph, and Noam Shazeer.
\newblock Switch transformers: Scaling to trillion parameter models with simple and efficient sparsity.
\newblock \emph{Journal of Machine Learning Research}, 23\penalty0 (120):\penalty0 1--39, 2022.

\bibitem[Gholami et~al.(2022)Gholami, Kim, Dong, Yao, Mahoney, and Keutzer]{gholami2022survey}
Amir Gholami, Sehoon Kim, Zhen Dong, Zhewei Yao, Michael~W Mahoney, and Kurt Keutzer.
\newblock A survey of quantization methods for efficient neural network inference.
\newblock In \emph{Low-power computer vision}, pp.\  291--326. Chapman and Hall/CRC, 2022.

\bibitem[Goyal et~al.(2017)Goyal, Khot, Summers-Stay, Batra, and Parikh]{goyal2017making}
Yash Goyal, Tejas Khot, Douglas Summers-Stay, Dhruv Batra, and Devi Parikh.
\newblock Making the v in vqa matter: Elevating the role of image understanding in visual question answering.
\newblock In \emph{Proceedings of the IEEE conference on computer vision and pattern recognition}, pp.\  6904--6913, 2017.

\bibitem[Gu et~al.(2023)Gu, Dong, Wei, and Huang]{gu2023knowledge}
Yuxian Gu, Li~Dong, Furu Wei, and Minlie Huang.
\newblock Knowledge distillation of large language models.
\newblock \emph{arXiv preprint arXiv:2306.08543}, 2023.

\bibitem[Gupta et~al.(2024)Gupta, Narasimhan, Jitkrittum, Rawat, Menon, and Kumar]{gupta2024language}
Neha Gupta, Harikrishna Narasimhan, Wittawat Jitkrittum, Ankit~Singh Rawat, Aditya~Krishna Menon, and Sanjiv Kumar.
\newblock Language model cascades: Token-level uncertainty and beyond.
\newblock \emph{arXiv preprint arXiv:2404.10136}, 2024.

\bibitem[Gupta \& Kembhavi(2023)Gupta and Kembhavi]{gupta2023visual}
Tanmay Gupta and Aniruddha Kembhavi.
\newblock Visual programming: Compositional visual reasoning without training.
\newblock In \emph{Proceedings of the IEEE/CVF Conference on Computer Vision and Pattern Recognition}, pp.\  14953--14962, 2023.

\bibitem[He et~al.(2016)He, Zhang, Ren, and Sun]{he2016deep}
Kaiming He, Xiangyu Zhang, Shaoqing Ren, and Jian Sun.
\newblock Deep residual learning for image recognition.
\newblock In \emph{Proceedings of the IEEE conference on computer vision and pattern recognition}, pp.\  770--778, 2016.

\bibitem[Hinton et~al.(2015)Hinton, Vinyals, and Dean]{hinton2015distilling}
Geoffrey Hinton, Oriol Vinyals, and Jeff Dean.
\newblock Distilling the knowledge in a neural network.
\newblock \emph{arXiv preprint arXiv:1503.02531}, 2015.

\bibitem[Hu et~al.(2024)Hu, Bieker, Li, Jiang, Keigwin, Ranganath, Keutzer, and Upadhyay]{hu2024routerbench}
Qitian~Jason Hu, Jacob Bieker, Xiuyu Li, Nan Jiang, Benjamin Keigwin, Gaurav Ranganath, Kurt Keutzer, and Shriyash~Kaustubh Upadhyay.
\newblock Routerbench: A benchmark for multi-llm routing system.
\newblock \emph{arXiv preprint arXiv:2403.12031}, 2024.

\bibitem[Huang et~al.(2024)Huang, An, Zhuang, Tao, Zhang, Jin, Xu, Chen, Huang, and Feng]{huang2024harder}
Quzhe Huang, Zhenwei An, Nan Zhuang, Mingxu Tao, Chen Zhang, Yang Jin, Kun Xu, Liwei Chen, Songfang Huang, and Yansong Feng.
\newblock Harder tasks need more experts: Dynamic routing in moe models.
\newblock \emph{arXiv preprint arXiv:2403.07652}, 2024.

\bibitem[Hudson \& Manning(2019)Hudson and Manning]{hudson2019gqa}
Drew~A Hudson and Christopher~D Manning.
\newblock Gqa: A new dataset for real-world visual reasoning and compositional question answering.
\newblock In \emph{Proceedings of the IEEE/CVF conference on computer vision and pattern recognition}, pp.\  6700--6709, 2019.

\bibitem[Hurst et~al.(2024)Hurst, Lerer, Goucher, Perelman, Ramesh, Clark, Ostrow, Welihinda, Hayes, Radford, et~al.]{hurst2024gpt}
Aaron Hurst, Adam Lerer, Adam~P Goucher, Adam Perelman, Aditya Ramesh, Aidan Clark, AJ~Ostrow, Akila Welihinda, Alan Hayes, Alec Radford, et~al.
\newblock Gpt-4o system card.
\newblock \emph{arXiv preprint arXiv:2410.21276}, 2024.

\bibitem[Jain et~al.(2021)Jain, Kothyari, Kumar, Jyothi, Ramakrishnan, and Chakrabarti]{jain2021select}
Aman Jain, Mayank Kothyari, Vishwajeet Kumar, Preethi Jyothi, Ganesh Ramakrishnan, and Soumen Chakrabarti.
\newblock Select, substitute, search: A new benchmark for knowledge-augmented visual question answering.
\newblock In \emph{Proceedings of the 44th International ACM SIGIR Conference on Research and Development in Information Retrieval}, pp.\  2491--2498, 2021.

\bibitem[Jitkrittum et~al.(2025)Jitkrittum, Narasimhan, Rawat, Juneja, Wang, Lee, Shenoy, Panigrahy, Menon, and Kumar]{jitkrittum2025universal}
Wittawat Jitkrittum, Harikrishna Narasimhan, Ankit~Singh Rawat, Jeevesh Juneja, Zifeng Wang, Chen-Yu Lee, Pradeep Shenoy, Rina Panigrahy, Aditya~Krishna Menon, and Sanjiv Kumar.
\newblock Universal model routing for efficient llm inference.
\newblock \emph{arXiv preprint arXiv:2502.08773}, 2025.

\bibitem[Johnson et~al.(2017)Johnson, Hariharan, Van Der~Maaten, Fei-Fei, Lawrence~Zitnick, and Girshick]{johnson2017clevr}
Justin Johnson, Bharath Hariharan, Laurens Van Der~Maaten, Li~Fei-Fei, C~Lawrence~Zitnick, and Ross Girshick.
\newblock Clevr: A diagnostic dataset for compositional language and elementary visual reasoning.
\newblock In \emph{Proceedings of the IEEE conference on computer vision and pattern recognition}, pp.\  2901--2910, 2017.

\bibitem[Khattab et~al.(2023)Khattab, Singhvi, Maheshwari, Zhang, Santhanam, Vardhamanan, Haq, Sharma, Joshi, Moazam, et~al.]{khattab2023dspy}
Omar Khattab, Arnav Singhvi, Paridhi Maheshwari, Zhiyuan Zhang, Keshav Santhanam, Sri Vardhamanan, Saiful Haq, Ashutosh Sharma, Thomas~T Joshi, Hanna Moazam, et~al.
\newblock Dspy: Compiling declarative language model calls into self-improving pipelines.
\newblock \emph{arXiv preprint arXiv:2310.03714}, 2023.

\bibitem[Kreikemeyer \& Andelfinger(2023)Kreikemeyer and Andelfinger]{kreikemeyer2023smoothing}
Justin~N Kreikemeyer and Philipp Andelfinger.
\newblock Smoothing methods for automatic differentiation across conditional branches.
\newblock \emph{IEEE Access}, 2023.

\bibitem[Kwon et~al.(2023)Kwon, Li, Zhuang, Sheng, Zheng, Yu, Gonzalez, Zhang, and Stoica]{kwon2023efficient}
Woosuk Kwon, Zhuohan Li, Siyuan Zhuang, Ying Sheng, Lianmin Zheng, Cody~Hao Yu, Joseph~E. Gonzalez, Hao Zhang, and Ion Stoica.
\newblock Efficient memory management for large language model serving with pagedattention.
\newblock In \emph{Proceedings of the ACM SIGOPS 29th Symposium on Operating Systems Principles}, 2023.

\bibitem[Leviathan et~al.(2023)Leviathan, Kalman, and Matias]{leviathan2023fast}
Yaniv Leviathan, Matan Kalman, and Yossi Matias.
\newblock Fast inference from transformers via speculative decoding.
\newblock In \emph{International Conference on Machine Learning}, pp.\  19274--19286. PMLR, 2023.

\bibitem[Li et~al.(2022{\natexlab{a}})Li, Li, Xiong, and Hoi]{li2022blip}
Junnan Li, Dongxu Li, Caiming Xiong, and Steven Hoi.
\newblock Blip: Bootstrapping language-image pre-training for unified vision-language understanding and generation.
\newblock In \emph{International conference on machine learning}, pp.\  12888--12900. PMLR, 2022{\natexlab{a}}.

\bibitem[Li et~al.(2023)Li, Li, Savarese, and Hoi]{li2023blip}
Junnan Li, Dongxu Li, Silvio Savarese, and Steven Hoi.
\newblock Blip-2: Bootstrapping language-image pre-training with frozen image encoders and large language models.
\newblock In \emph{International conference on machine learning}, pp.\  19730--19742. PMLR, 2023.

\bibitem[Li et~al.(2022{\natexlab{b}})Li, Zhang, Zhang, Yang, Li, Zhong, Wang, Yuan, Zhang, Hwang, et~al.]{li2022grounded}
Liunian~Harold Li, Pengchuan Zhang, Haotian Zhang, Jianwei Yang, Chunyuan Li, Yiwu Zhong, Lijuan Wang, Lu~Yuan, Lei Zhang, Jenq-Neng Hwang, et~al.
\newblock Grounded language-image pre-training.
\newblock In \emph{Proceedings of the IEEE/CVF Conference on Computer Vision and Pattern Recognition}, pp.\  10965--10975, 2022{\natexlab{b}}.

\bibitem[Li(2025)]{li2025llm}
Yang Li.
\newblock Llm bandit: Cost-efficient llm generation via preference-conditioned dynamic routing.
\newblock \emph{arXiv preprint arXiv:2502.02743}, 2025.

\bibitem[Lin et~al.(2024)Lin, Tang, Tang, Yang, Chen, Wang, Xiao, Dang, Gan, and Han]{lin2024awq}
Ji~Lin, Jiaming Tang, Haotian Tang, Shang Yang, Wei-Ming Chen, Wei-Chen Wang, Guangxuan Xiao, Xingyu Dang, Chuang Gan, and Song Han.
\newblock Awq: Activation-aware weight quantization for on-device llm compression and acceleration.
\newblock \emph{Proceedings of Machine Learning and Systems}, 6:\penalty0 87--100, 2024.

\bibitem[Lin et~al.(2014)Lin, Maire, Belongie, Hays, Perona, Ramanan, Doll{\'a}r, and Zitnick]{lin2014microsoft}
Tsung-Yi Lin, Michael Maire, Serge Belongie, James Hays, Pietro Perona, Deva Ramanan, Piotr Doll{\'a}r, and C~Lawrence Zitnick.
\newblock Microsoft coco: Common objects in context.
\newblock In \emph{Computer Vision--ECCV 2014: 13th European Conference, Zurich, Switzerland, September 6-12, 2014, Proceedings, Part V 13}, pp.\  740--755. Springer, 2014.

\bibitem[Liu et~al.(2024)Liu, Li, Li, and Lee]{liu2024improved}
Haotian Liu, Chunyuan Li, Yuheng Li, and Yong~Jae Lee.
\newblock Improved baselines with visual instruction tuning.
\newblock In \emph{Proceedings of the IEEE/CVF Conference on Computer Vision and Pattern Recognition}, pp.\  26296--26306, 2024.

\bibitem[Liu et~al.(2023)Liu, Zeng, Ren, Li, Zhang, Yang, Li, Yang, Su, Zhu, et~al.]{liu2023grounding}
Shilong Liu, Zhaoyang Zeng, Tianhe Ren, Feng Li, Hao Zhang, Jie Yang, Chunyuan Li, Jianwei Yang, Hang Su, Jun Zhu, et~al.
\newblock Grounding dino: Marrying dino with grounded pre-training for open-set object detection.
\newblock \emph{arXiv preprint arXiv:2303.05499}, 2023.

\bibitem[Lu et~al.(2023)Lu, Yuan, Lin, Lin, Yuan, Zhou, and Zhou]{lu2023routing}
Keming Lu, Hongyi Yuan, Runji Lin, Junyang Lin, Zheng Yuan, Chang Zhou, and Jingren Zhou.
\newblock Routing to the expert: Efficient reward-guided ensemble of large language models.
\newblock \emph{arXiv preprint arXiv:2311.08692}, 2023.

\bibitem[Lu et~al.(2024)Lu, Peng, Cheng, Galley, Chang, Wu, Zhu, and Gao]{lu2024chameleon}
Pan Lu, Baolin Peng, Hao Cheng, Michel Galley, Kai-Wei Chang, Ying~Nian Wu, Song-Chun Zhu, and Jianfeng Gao.
\newblock Chameleon: Plug-and-play compositional reasoning with large language models.
\newblock \emph{Advances in Neural Information Processing Systems}, 36, 2024.

\bibitem[Miao et~al.(2023)Miao, Oliaro, Zhang, Cheng, Wang, Zhang, Wong, Zhu, Yang, Shi, et~al.]{miao2023specinfer}
Xupeng Miao, Gabriele Oliaro, Zhihao Zhang, Xinhao Cheng, Zeyu Wang, Zhengxin Zhang, Rae Ying~Yee Wong, Alan Zhu, Lijie Yang, Xiaoxiang Shi, et~al.
\newblock Specinfer: Accelerating generative large language model serving with tree-based speculative inference and verification.
\newblock \emph{arXiv preprint arXiv:2305.09781}, 2023.

\bibitem[Nguyen et~al.(2024)Nguyen, Hoang, Decugis, Manchanda, Chawla, and Doan]{nguyen2024metallm}
Quang~H Nguyen, Duy~C Hoang, Juliette Decugis, Saurav Manchanda, Nitesh~V Chawla, and Khoa~D Doan.
\newblock Metallm: A high-performant and cost-efficient dynamic framework for wrapping llms.
\newblock \emph{arXiv preprint arXiv:2407.10834}, 2024.

\bibitem[Nie et~al.(2024)Nie, Ding, Hu, Jermaine, and Chaudhuri]{nie2024online}
Lunyiu Nie, Zhimin Ding, Erdong Hu, Christopher Jermaine, and Swarat Chaudhuri.
\newblock Online cascade learning for efficient inference over streams.
\newblock In \emph{Forty-first International Conference on Machine Learning}, 2024.

\bibitem[{OpenAI}(2024)]{openai20244o}
{OpenAI}.
\newblock Gpt-4o mini: Advancing cost-efficient intelligence.
\newblock \url{https://openai. com/index/gpt-4o-mini-advancing-cost-efficient-intelligence}, 2024.

\bibitem[Schwenk et~al.(2022)Schwenk, Khandelwal, Clark, Marino, and Mottaghi]{schwenk2022okvqa}
Dustin Schwenk, Apoorv Khandelwal, Christopher Clark, Kenneth Marino, and Roozbeh Mottaghi.
\newblock A-okvqa: A benchmark for visual question answering using world knowledge.
\newblock In \emph{European conference on computer vision}, pp.\  146--162. Springer, 2022.

\bibitem[Shnitzer et~al.(2023)Shnitzer, Ou, Silva, Soule, Sun, Solomon, Thompson, and Yurochkin]{shnitzer2023large}
Tal Shnitzer, Anthony Ou, M{\'\i}rian Silva, Kate Soule, Yuekai Sun, Justin Solomon, Neil Thompson, and Mikhail Yurochkin.
\newblock Large language model routing with benchmark datasets.
\newblock \emph{arXiv preprint arXiv:2309.15789}, 2023.

\bibitem[Subramanian et~al.(2023)Subramanian, Narasimhan, Khangaonkar, Yang, Nagrani, Schmid, Zeng, Darrell, and Klein]{subramanian2023modular}
Sanjay Subramanian, Medhini Narasimhan, Kushal Khangaonkar, Kevin Yang, Arsha Nagrani, Cordelia Schmid, Andy Zeng, Trevor Darrell, and Dan Klein.
\newblock Modular visual question answering via code generation.
\newblock \emph{arXiv preprint arXiv:2306.05392}, 2023.

\bibitem[Sur{\'\i}s et~al.(2023)Sur{\'\i}s, Menon, and Vondrick]{suris2023vipergpt}
D{\'\i}dac Sur{\'\i}s, Sachit Menon, and Carl Vondrick.
\newblock Vipergpt: Visual inference via python execution for reasoning.
\newblock In \emph{Proceedings of the IEEE/CVF International Conference on Computer Vision}, pp.\  11888--11898, 2023.

\bibitem[Wang et~al.(2022)Wang, Yang, Men, Lin, Bai, Li, Ma, Zhou, Zhou, and Yang]{wang2022ofa}
Peng Wang, An~Yang, Rui Men, Junyang Lin, Shuai Bai, Zhikang Li, Jianxin Ma, Chang Zhou, Jingren Zhou, and Hongxia Yang.
\newblock Ofa: Unifying architectures, tasks, and modalities through a simple sequence-to-sequence learning framework.
\newblock In \emph{International conference on machine learning}, pp.\  23318--23340. PMLR, 2022.

\bibitem[Williams(1992)]{williams1992simple}
Ronald~J Williams.
\newblock Simple statistical gradient-following algorithms for connectionist reinforcement learning.
\newblock \emph{Machine learning}, 8:\penalty0 229--256, 1992.

\bibitem[Wu et~al.(2023)Wu, Bansal, Zhang, Wu, Zhang, Zhu, Li, Jiang, Zhang, and Wang]{wu2023autogen}
Qingyun Wu, Gagan Bansal, Jieyu Zhang, Yiran Wu, Shaokun Zhang, Erkang Zhu, Beibin Li, Li~Jiang, Xiaoyun Zhang, and Chi Wang.
\newblock Autogen: Enabling next-gen llm applications via multi-agent conversation framework.
\newblock \emph{arXiv preprint arXiv:2308.08155}, 2023.

\bibitem[Xu et~al.(2024)Xu, Yin, Cai, Yi, Xu, Wang, Wu, Zhao, Yang, Wang, et~al.]{xu2024survey}
Mengwei Xu, Wangsong Yin, Dongqi Cai, Rongjie Yi, Daliang Xu, Qipeng Wang, Bingyang Wu, Yihao Zhao, Chen Yang, Shihe Wang, et~al.
\newblock A survey of resource-efficient llm and multimodal foundation models.
\newblock \emph{arXiv preprint arXiv:2401.08092}, 2024.

\bibitem[Zhang et~al.(2016)Zhang, Goyal, Summers-Stay, Batra, and Parikh]{zhang2016yin}
Peng Zhang, Yash Goyal, Douglas Summers-Stay, Dhruv Batra, and Devi Parikh.
\newblock Yin and yang: Balancing and answering binary visual questions.
\newblock In \emph{Proceedings of the IEEE conference on computer vision and pattern recognition}, pp.\  5014--5022, 2016.

\bibitem[Zhang et~al.(2020)Zhang, Zhou, Li, and Gu]{zhang2020neural}
Weitong Zhang, Dongruo Zhou, Lihong Li, and Quanquan Gu.
\newblock Neural thompson sampling.
\newblock \emph{arXiv preprint arXiv:2010.00827}, 2020.

\bibitem[Zinkevich(2003)]{zinkevich2003online}
Martin Zinkevich.
\newblock Online convex programming and generalized infinitesimal gradient ascent.
\newblock In \emph{Proceedings of the 20th international conference on machine learning (icml-03)}, pp.\  928--936, 2003.

\end{thebibliography}
\bibliographystyle{colm2025_conference}

\appendix
\newtheorem{theorem}{Theorem}
\newtheorem{assumption}{Assumption}

\section{No-Regret Guarantee for Structured REINFORCE}
\label{sec:app_proof}
We establish that our proposed structured REINFORCE algorithm achieves a no-regret guarantee in an online learning setting. Before presenting the main theorem, we outline the key assumptions that underpin our analysis:

\begin{assumption}[Bounded Rewards]
For all time steps $t$, configurations $\vec{v}_t \in V$, and inputs $x_t \in \mathcal{D}$, the reward satisfies $R(\vec{v}_t, x_t, y_t) \in [R_{\min}, R_{\max}]$, where $R_{\max} - R_{\min} < \infty$.
\end{assumption}

\begin{assumption}[Policy Expressiveness]
For each input $x_t$, there exists an optimal configuration $\vec{v}^*_{x_t} \in V$ that maximizes $R(\vec{v}, x_t, y_t)$. Moreover, the policy class is sufficiently expressive such that there exist parameters $\{\theta_{k_i}^*\}_{i=1}^N$ for which $\pi_{k_i}(m_{k_i,j_i}^* \mid x_t; \theta_{k_i}^*) \approx 1$, where $m_{k_i,j_i}^*$ is the optimal backend for $f_{k_i}$ in $\vec{v}^*_{x_t}$.
\end{assumption}

\begin{assumption}[Sufficient Exploration]
The algorithm employs Thompson Sampling with an exploration parameter $\nu > 0$, ensuring that every backend $m_{k_i, j_i} \in M_{k_i}$ has a non-zero probability of being sampled at each time step $t$.
\end{assumption}

\begin{assumption}[Convergence of Policy Gradient]
The learning rate $\eta_t$ is set to $1/\sqrt{t}$, and the policy parameterization (e.g., softmax over $M_{k_i}$) guarantees that gradient updates converge to a near-optimal policy \cite{agarwal2021theory}.
\end{assumption}

\begin{assumption}[Stationary Input Distribution]
Inputs $x_t$ are drawn independently and identically distributed (i.i.d.) from a fixed distribution $\mathcal{D}$, ensuring a consistent optimal policy over time.
\end{assumption}

With these assumptions in place, we can formally state the main result:

\begin{theorem}
Under Assumptions 1--5, the structured REINFORCE algorithm is no-regret, meaning that the average regret satisfies:
\begin{equation*}
\frac{\gamma_T}{T} \to 0 \quad \text{as} \quad T \to \infty,
\end{equation*}
both in expectation and with high probability, where the regret $\gamma_T$ is defined as:
\begin{equation*}
\gamma_T = \max_{\vec{v} \in V} \sum_{t=1}^T R(\vec{v}, x_t, y_t) - \sum_{t=1}^T R(\vec{v}_t, x_t, y_t).
\end{equation*}
\end{theorem}

We now prove this theorem, showing that the algorithm’s regret diminishes over time. The proof proceeds by defining the regret, analyzing the algorithm’s behavior under the assumptions, and bounding the regret both in expectation and with high probability.

\subsection*{Proof of Theorem 1}
We start with giving the formal definition of regret $\gamma_T$, which measures the cumulative difference between the maximum achievable reward and the algorithm’s actual reward over $T$ steps:
\begin{equation*}
\gamma_T = \max_{\vec{v} \in V} \sum_{t=1}^T R(\vec{v}, x_t, y_t) - \sum_{t=1}^T R(\vec{v}_t, x_t, y_t).
\end{equation*}
Our goal is to show that the average regret, $\frac{\gamma_T}{T}$, approaches zero as $T \to \infty$. To simplify the analysis, we use a stronger benchmark: the optimal context-dependent configuration $\vec{v}^*_{x_t}$ that maximizes $R(\vec{v}, x_t, y_t)$ for each $x_t$. Since $\max_{\vec{v} \in V} R(\vec{v}, x_t, y_t) \leq R(\vec{v}^*_{x_t}, x_t, y_t)$, we have:
\begin{equation*}
\gamma_T \leq \sum_{t=1}^T R(\vec{v}^*_{x_t}, x_t, y_t) - \sum_{t=1}^T R(\vec{v}_t, x_t, y_t).
\end{equation*}
This upper bound focuses the proof on the gap between the optimal and achieved rewards per step.

The structured REINFORCE algorithm updates sub-policy parameters $\theta_{k_i}$ using policy gradients. The expected cumulative reward is:
\begin{equation*}
\mathcal{J}(\pi) = \mathbb{E}_{\pi} \left[ \sum_{t=1}^T R(\vec{v}_t, x_t, y_t) \right],
\end{equation*}
with the gradient for each sub-policy:
\begin{equation*}
\nabla_{\theta_{k_i}} \mathcal{J}(\pi_{k_i}) = \sum_{t=1}^T \mathbb{E}_{j_i^* \sim \pi_{k_i}(\cdot \mid x_t)} \left[ \nabla_{\theta_{k_i}} \log \pi_{k_i}(m_{k_i,j_i^*} \mid x_t; \theta_{k_i}) \cdot r_{k_i,j_i^*} \right].
\end{equation*}
The algorithm approximates this gradient with a single sample, updating parameters as:
\begin{equation*}
\theta_{k_i, t+1} = \theta_{k_i, t} + \eta_t \nabla_{\theta_{k_i}} \log \pi_{k_i}(m_{k_i,j_i^*} \mid x_t; \theta_{k_i, t}) r_{k_i,j_i^*},
\end{equation*}
where $\eta_t = 1/\sqrt{t}$ (Assumption 4), and the sub-reward is defined as:
\begin{equation*}
r_{k_i,j_i^*} = -\lambda \mathcal{C}(m_{k_i,j_i^*}) - \frac{1}{N} \mathcal{L}(p(x_t \mid \vec{v}_t), y_t).
\end{equation*}
Assumptions 2 (expressive policy class) and 3 (sufficient exploration via Thompson Sampling) ensure that each sub-policy $\pi_{k_i}$ converges to the optimal sub-policy $\pi_{k_i}^*$, where $\pi_{k_i}^*(m_{k_i,j_i}^* \mid x_t) \approx 1$ for the optimal backend $m_{k_i,j_i}^*$ in $\vec{v}^*_{x_t}$. Thus, the overall policy $\pi_t$ converges to the optimal policy $\pi^*$, leading to $\mathbb{E}_{\pi_t} [R(\vec{v}_t, x_t, y_t)] \to R(\vec{v}^*_{x_t}, x_t, y_t)$ as $t \to \infty$.

Define the per-step regret as:
\begin{equation*}
\rho_t = R(\vec{v}^*_{x_t}, x_t, y_t) - R(\vec{v}_t, x_t, y_t),
\end{equation*}
with expected value:
\begin{equation*}
\mathbb{E}[\rho_t] = \mathbb{E}_{x_t \sim \mathcal{D}} \left[ R(\vec{v}^*_{x_t}, x_t, y_t) - \mathbb{E}_{\vec{v}_t \sim \pi_t} [R(\vec{v}_t, x_t, y_t)] \right].
\end{equation*}
Since $\pi_t \to \pi^*$, we have $\mathbb{E}[\rho_t] \to 0$. The total expected regret is:
\begin{equation*}
\mathbb{E}[\gamma_T] \leq \sum_{t=1}^T \mathbb{E}[\rho_t].
\end{equation*}
Given $\eta_t = 1/\sqrt{t}$ and standard policy gradient convergence (Assumption 4), we bound:
\begin{equation*}
\mathbb{E}[\gamma_T] \leq C \sqrt{T},
\end{equation*}
for some constant $C$ based on the reward bounds and policy parameters \cite{zinkevich2003online, ban2021multi}. Thus, the average expected regret satisfies:
\begin{equation*}
\frac{\mathbb{E}[\gamma_T]}{T} \leq \frac{C}{\sqrt{T}} \to 0 \quad \text{as} \quad T \to \infty,
\end{equation*}
establishing no-regret in expectation.

Now, we extend this to show that the actual regret $\gamma_T$ converges similarly with high probability. By Assumption 1, rewards are bounded, so $\rho_t \in [-(R_{\max} - R_{\min}), R_{\max} - R_{\min}]$, and let $B = R_{\max} - R_{\min}$. Since $x_t$ are i.i.d. (Assumption 5) and $\vec{v}_t$ are sampled independently given $x_t$ and $\pi_t$, the $\rho_t$ are independent. Applying Hoeffding’s inequality to $\gamma_T = \sum_{t=1}^T \rho_t$:
\begin{equation*}
\mathbb{P}\left( \left| \gamma_T - \mathbb{E}[\gamma_T] \right| \geq \epsilon \right) \leq 2 \exp\left( -\frac{2 \epsilon^2}{T B^2} \right).
\end{equation*}
Set $\epsilon = \delta T$, so:
\begin{equation*}
\mathbb{P}\left( \left| \gamma_T - \mathbb{E}[\gamma_T] \right| \geq \delta T \right) \leq 2 \exp\left( -\frac{2 \delta^2 T}{B^2} \right).
\end{equation*}
This probability approaches 0 exponentially as $T \to \infty$. Thus, with probability at least $1 - 2 \exp\left( -\frac{2 \delta^2 T}{B^2} \right)$, which approaches 1 as $T$ grows, we have:
\begin{equation*}
\gamma_T < \mathbb{E}[\gamma_T] + \delta T \leq C \sqrt{T} + \delta T.
\end{equation*}
Dividing by $T$, we find:
\begin{equation*}
\frac{\gamma_T}{T} < \frac{C}{\sqrt{T}} + \delta.
\end{equation*}
For any $\epsilon > 0$, choose $\delta = \frac{\epsilon}{2}$ and $T > \left( \frac{2C}{\epsilon} \right)^2$, so $\frac{C}{\sqrt{T}} < \frac{\epsilon}{2}$, yielding:
\begin{equation*}
\frac{\gamma_T}{T} < \epsilon,
\end{equation*}
with probability approaching 1. Hence, $\frac{\gamma_T}{T} \to 0$ with high probability, completing the no-regret proof.
\section{Detailed Experimental Setups}
All experiments are conducted on a single machine equipped with 8 NVIDIA A40 GPUs (48GB memory each), running CUDA 12.4. Detailed package versions are listed in the environment file available at \href{https://github.com/Flitternie/FMProgramming/blob/master/environment.yml}{\texttt{GitHub}}. Hyperparameter settings are specified in the experimental configuration files for \href{https://github.com/Flitternie/FMProgramming/blob/master/config/binary_vqa.yaml}{\texttt{binary VQA}} and 
\href{https://github.com/Flitternie/FMProgramming/blob/master/config/open_form_vqa.yaml}{\texttt{open-form VQA}}.

For the offline code generation, we use a prompt format similar to ViperGPT, including the DSL, the user query, and a one-shot example. In practice, the generated programs are $>$90\% correct, and errors are easily caught during dry runs or audits. A full prompt example is also available at \href{https://github.com/Flitternie/FMProgramming/blob/master/dsl.prompt}{\texttt{GitHub}}.

The current FM backend is set up with VLLM \cite{kwon2023efficient}, Huggingface Hub, MMDetection \cite{chen2019mmdetection}, and AgentLego \cite{contributors7agentlego}. Our framework supports a flexible backend construction with any open-source or closed-source API-based models. However, due to the unavailability of computational costs of closed-source models, such as GPT and Gemini series models, we do not include them in our FM backend during experiments. 

For open-form VQA experiments, we consistently use the same system prompt for MLLMs:

\begin{exmp}{}{demo}
\texttt{Keep your answer short. Try to answer within 3 words. 
For numerical answers, use number digits (\eg 5 instead of five), and returns the number only. 
If there are multiple answers, separate them with a comma (\eg cat, dog).
If you find the question unanswerable based on the image content, output "N/A". 
For example, if the image content is irrelevant to the question, or the content in the image does not fully and clearly match all the entities, humans, attributes, spatial, logical, and numerical constraints in the question, output "N/A".}
\end{exmp}

\begin{figure*}[t]
     \centering
     \includegraphics[width=\textwidth]{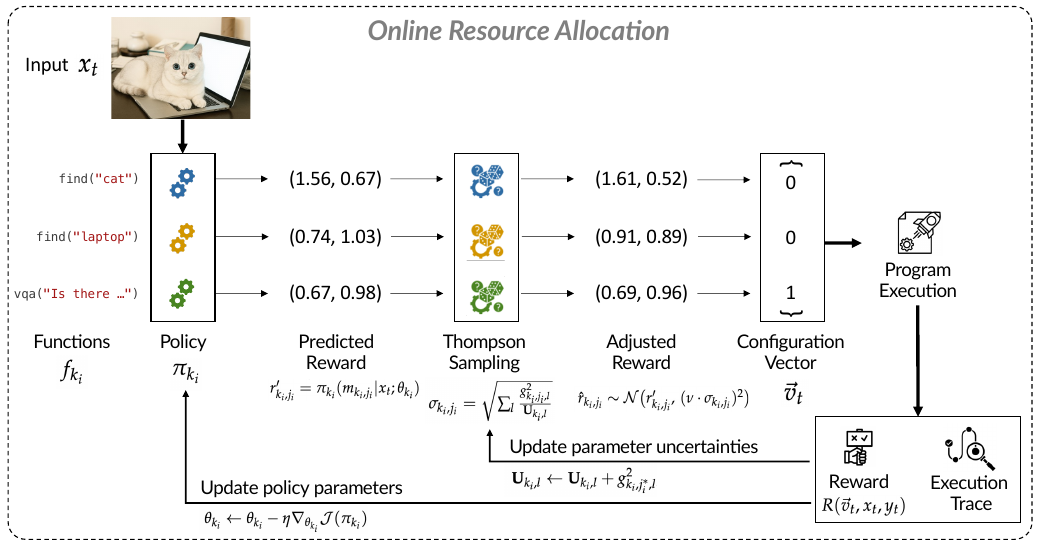}
     \caption{A walk-through example showing how an FMP is compiled and routed with online resource allocation in Algorithm \ref{algo:structured_reinforce}. Here we look at a specific input $x_t$ for illustration but the user inputs actually arrive continuously in a stream. } 
     \label{fig:demo}
\end{figure*}

\section{Detailed Benchmark Construction and Evaluation}

\subsection{Streaming Binary VQA}
\label{sec:appendix_benchmarks_a}
\begin{figure*}[h]
     \centering
     \includegraphics[width=.9\textwidth]{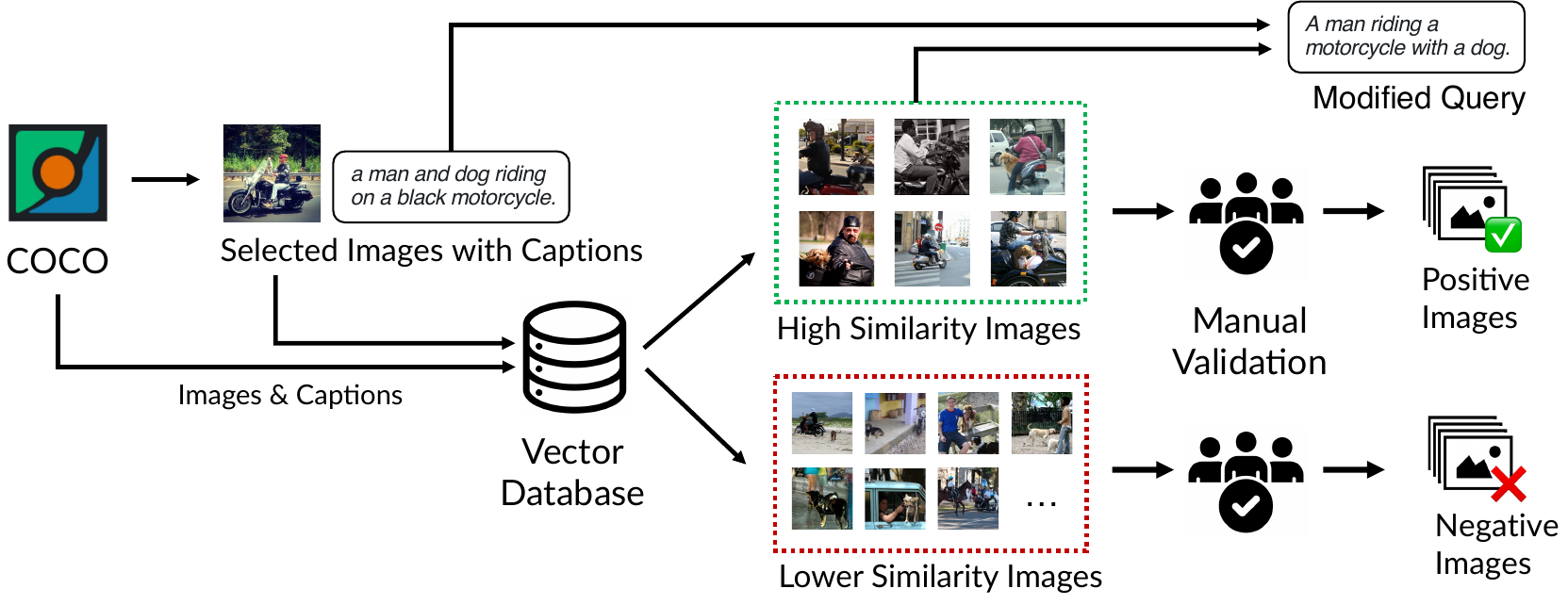}
     \caption{Benchmark construction pipeline for the Streaming Binary VQA dataset. } 
     \label{fig:retrieval}
\end{figure*}

\paragraph{Benchmark Construction} Our dataset is constructed from COCO \cite{lin2014microsoft}, selecting captions that require multi-object compositional reasoning with spatial, logical, or numerical constraints. For each query, we prepare a set of more than 2,000 images, sampled based on the similarity of their captions to the query. The system must output a binary decision (\textit{yes/no}) for each image indicating whether it satisfies the compositional query. To enforce structured reasoning, we leverage an LLM\footnote{We use \texttt{GPT-4o} \cite{hurst2024gpt} as the LLM throughout this section unless otherwise specified.} to generate FM programs in a predefined DSL. These programs decompose the query into discrete reasoning steps, guiding the selection of foundation models for subtask execution. A detailed pipeline for benchmark construction is illustrated in Figure~\ref{fig:retrieval}.

\paragraph{Annotation and Verification} Eleven human annotators validate the correctness of the neurosymbolic programs and the image labels, ensuring that each program aligns with the intended reasoning process and each image is correctly labeled against the compositional constraints. The final dataset consists of 33 queries covering three primary reasoning types (note that a query may fall into multiple types):
\begin{itemize}[leftmargin=0.15in, itemsep=0em]
    \item \textbf{Spatial Reasoning} (20 queries): These queries require understanding and interpreting the spatial relationships between objects or people in an image. They often describe where things are located relative to each other, \eg “\textit{Is there a person riding a bicycle next to a bus on the street}”.
    \item \textbf{Logical Reasoning} (15 queries): These queries involve conditions, attributes, or combinations that require deductive thinking. The model needs to process logical relationships, such as inclusion, exclusion, or conjunction, \eg “\textit{Are there people riding bikes, scooters, or motorcycles while holding or using umbrellas?}”.
    \item \textbf{Numerical Reasoning} (9 queries): These queries test the ability to understand and count quantities or numbers in a scene. They often specify exact counts or comparisons, \eg “\textit{Are there at least four horses on a beach}”.
\end{itemize}
\paragraph{Evaluation} Since the benchmark is highly imbalanced, with a positive-to-negative ratio of around 1:100, task performance is measured using accuracy, recall, precision, and F1-score.

\subsection{Streaming Open-form VQA}
\label{sec:appendix_benchmarks_b}
\paragraph{Benchmark Construction} To enable evaluation on more complex open-form questions, we construct a dataset comprising 50 queries and 25,000 images, spanning five distinct reasoning categories. To ensure the validity and diversity of these queries, we randomly sample them from established benchmark datasets. Specifically, spatial queries are drawn from the GQA dataset \cite{hudson2019gqa}, focusing on queries labeled as \texttt{relS} and \texttt{categoryRelS}. Logical queries are also sampled from GQA, targeting the detailed types \texttt{twoCommon}, \texttt{twoSameMaterialC}, \texttt{twoDifferentC}, and \texttt{twoDifferent}. Numerical queries are selected from the A-OKVQA dataset \cite{schwenk2022okvqa}, while comparative and external knowledge queries are sourced from OKVQAS3 \cite{jain2021select}. To enhance clarity and naturalness, some of the queries are manually rewritten.

Each query is associated with 500 generated images. The image generation pipeline begins with an LLM generating a set of 10 possible answers for each query, proposing potential scene setups along with 3 additional objects, and constructing detailed image descriptions. These descriptions are then used to prompt a diffusion model\footnote{We use \texttt{FLUX.1-dev} \cite{flux2023} as the diffusion model for image generation.} for image generation. 

To assess model robustness and reasoning precision, we incorporate unanswerable images that are visually coherent but semantically invalid with respect to the query. These include both unrelated (random) images and images that are intentionally crafted to closely resemble answerable cases, making them more difficult to distinguish, as illustrated in Figure \ref{fig:example}. This setup challenges models not only to infer the correct answer when possible but also to recognize when a question cannot be answered from the image. 

Additionally, for each query, we also synthesize the corresponding FM program in the predefined DSL, providing a structured decomposition of the reasoning process.

\begin{figure*}[h]
     \centering
     \includegraphics[width=\textwidth]{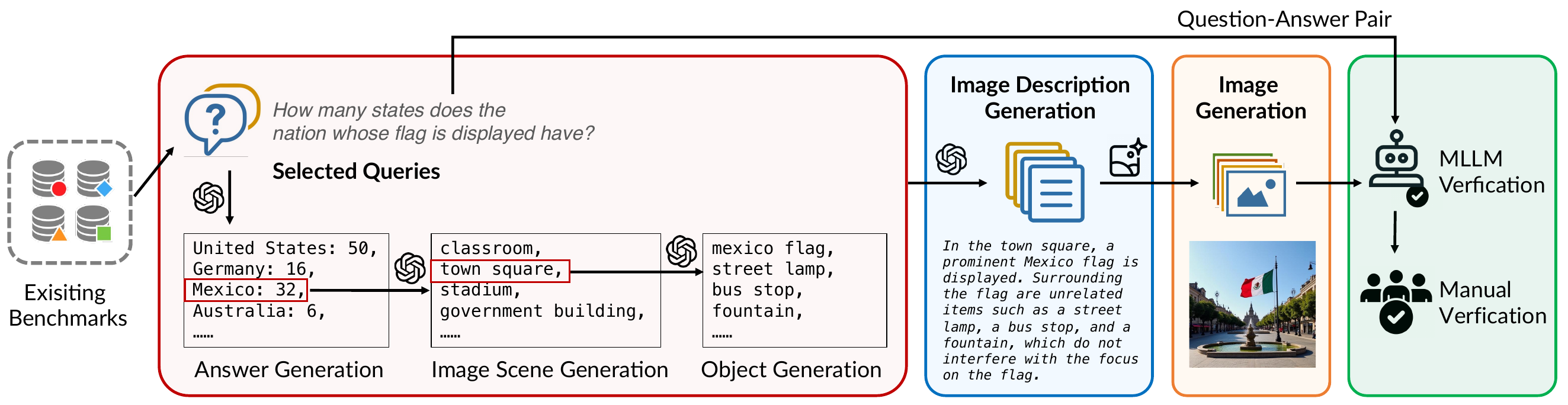}
     \caption{Benchmark construction pipeline for the Streaming Open-form VQA dataset.} 
     \label{fig:vqa}
\end{figure*}

\begin{figure*}[t]
    \centering
    \begin{subfigure}[t]{.3\textwidth}
        \centering
        \includegraphics[height=1.75in]{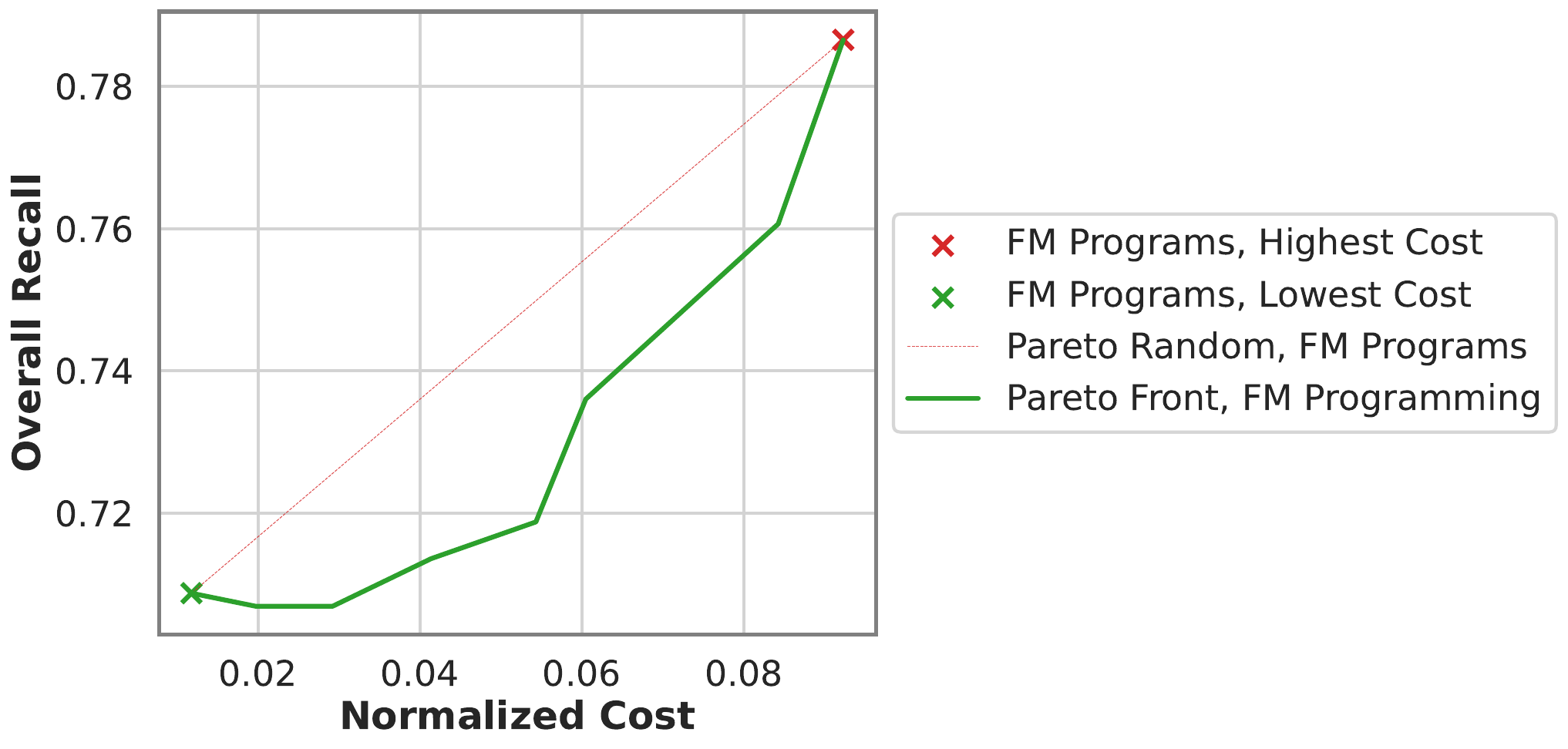}
    \end{subfigure}%
    ~ 
    \begin{subfigure}[t]{.8\textwidth}
        \centering
        \includegraphics[height=1.75in]{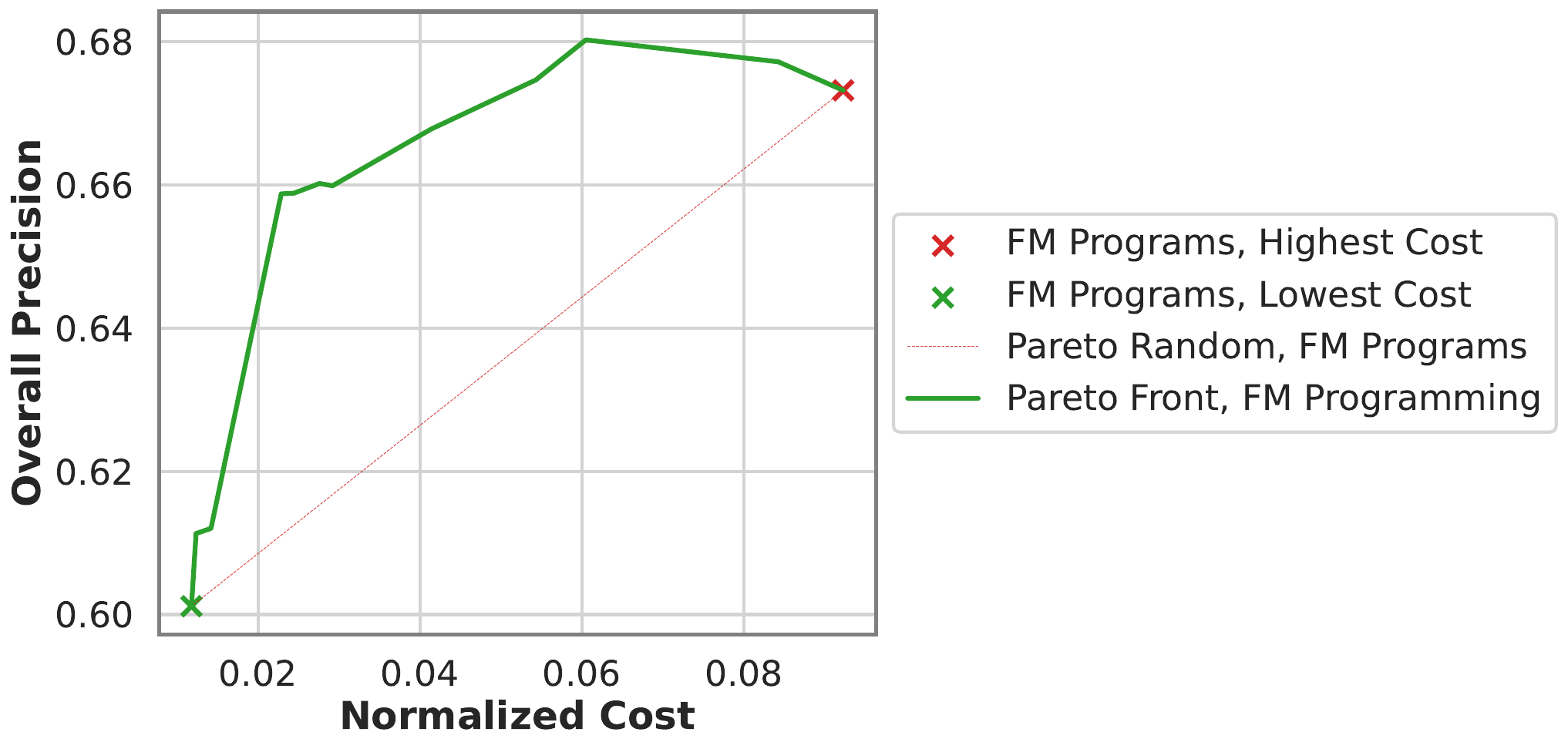}
    \end{subfigure}
\caption{Pareto front of precision and recall scores on the Streaming Binary VQA benchmark. The policy prioritizes precision over recall as the cost budget increases. }
\label{fig:verification_recall_precision}
\end{figure*}

\begin{figure*}[t]
    \centering
    \begin{subfigure}[t]{.3\textwidth}
        \centering
        \includegraphics[height=1.8in]{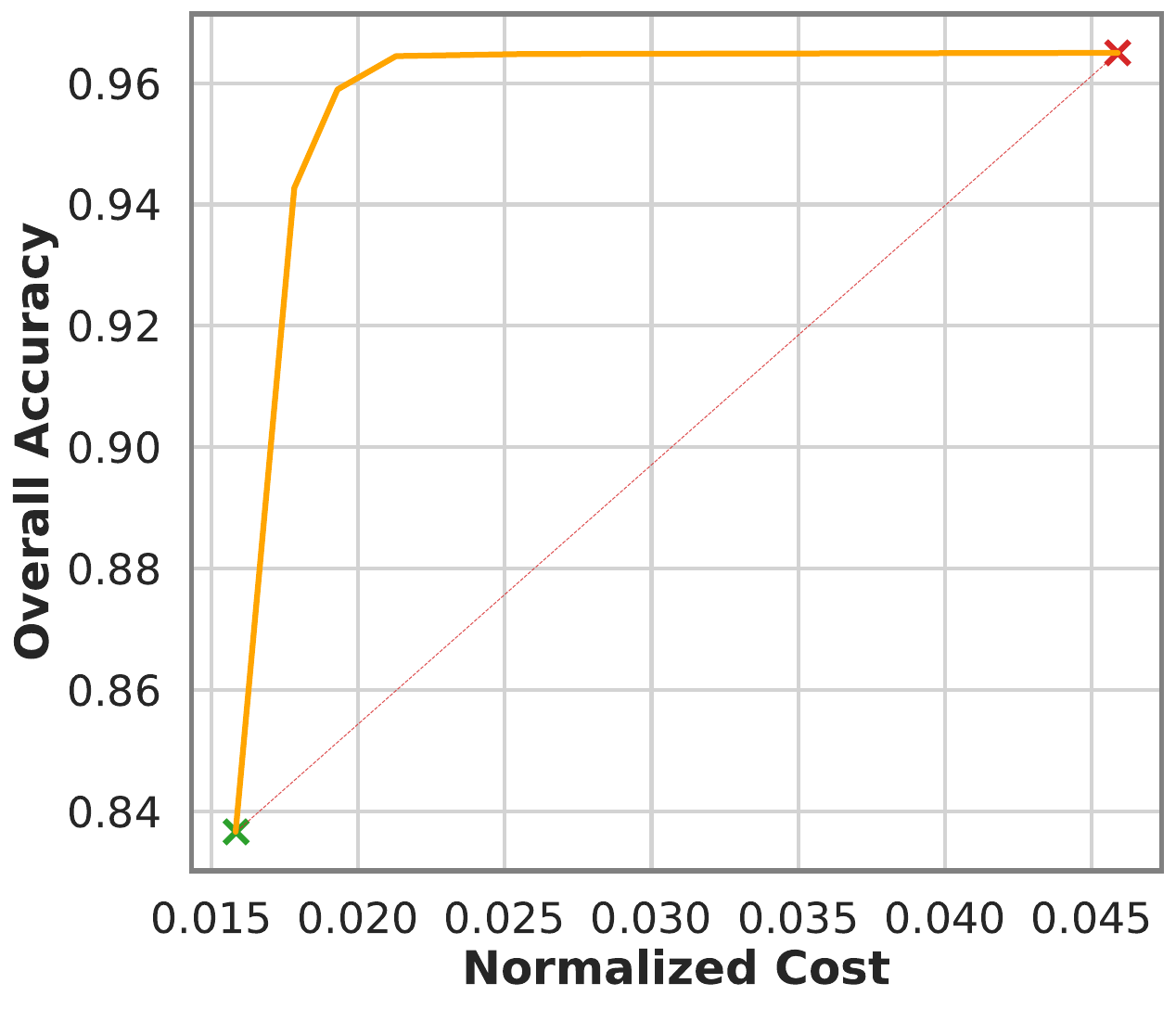}
    \end{subfigure}%
    ~ 
    \begin{subfigure}[t]{.8\textwidth}
        \centering
        \includegraphics[height=1.8in]{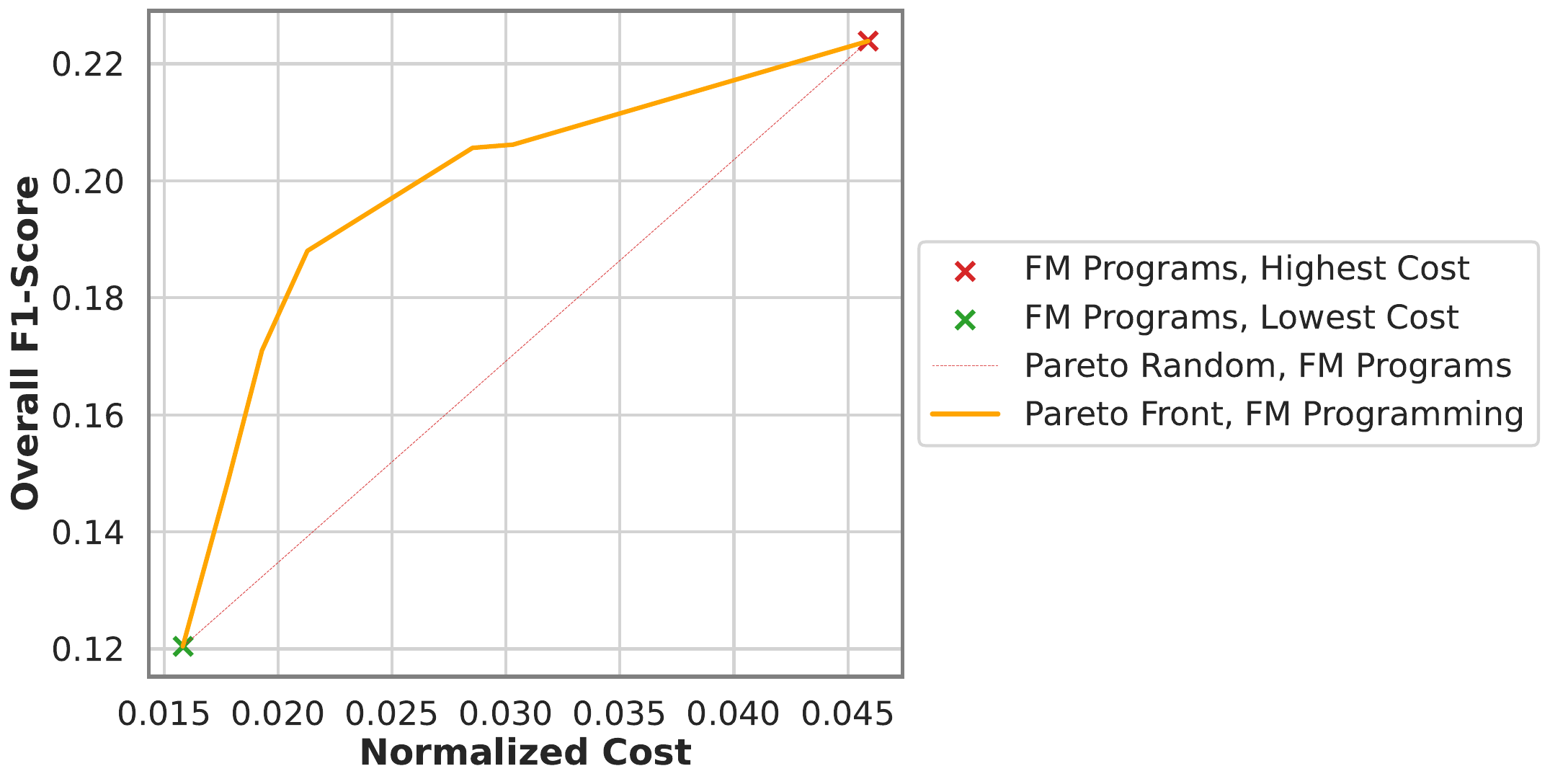}
    \end{subfigure}
    \caption{Additional results on the Streaming Binary VQA benchmark with different FM backends. Costs are normalized based on the inference costs of Qwen2.5-VL 72B. The backends include GLIP tiny (231M) and base (430M) models \cite{li2022grounded} for object detection, OFA base (182M) \cite{wang2022ofa} and BLIP-2 OPT-2.7B (3.745B) \cite{li2023blip} for language-vision understanding. FM programming outperforms the baseline with better cost-performance trade-offs, demonstrating its adaptability across backend setups. }
    \label{fig:verification_add}
\end{figure*}

\begin{table*}[t]
    \centering
    \resizebox{\textwidth}{!}{
    \begin{tabular}{llcccc|c|c}
        \toprule
        & & \multirow{2}{*}{VQA  v2.0} &  \multirow{2}{*}{GQA} & \multirow{2}{*}{CLEVR} & \multirow{2}{*}{A-OKVQA} & \multicolumn{2}{c}{\textbf{Streaming VQA}} \\
        & & & & & & Binary & Open-form\\
        \midrule
        \multirow{6}{*}{\textbf{Query}} 
        & Multiple Objects & \xmark & \cmark & \cmark & \cmark & \cmark & \cmark \\
        & Spatial Reasoning & \cmark & \cmark & \cmark & \cmark & \cmark & \cmark\\
        & Logical Reasoning & \cmark & \cmark & \cmark & \cmark & \cmark & \cmark\\
        & Numerical Reasoning & \cmark & \cmark & \cmark & \cmark & \cmark & \cmark\\
        & Comparative Reasoning & \cmark & \cmark & \cmark & \cmark & \xmark & \cmark\\
        & External Knowledge & \xmark & \xmark & \xmark & \cmark & \xmark & \cmark\\
        \midrule
        \multirow{2}{*}{\textbf{Image}} 
        & Source & COCO & COCO \& Flickr & Synthetic & COCO & COCO & Generation \\
        & Unanswerable Images & \xmark & \xmark & \xmark & \xmark & \xmark & \cmark\\
        \midrule
        \multirow{3}{*}{\textbf{Scale}} 
        & \# Queries & 1.1M & 22M & 999,968 & 24,903 & 33 & 50 \\
        & \# Images & 200K & 113K & 100,000 & 23,692 & 66,279 & 25,000 \\
        & \textbf{\# Image(s) per query}  & 2 & 1 & 1 & 1 & $>$\textbf{2000} & \textbf{500} \\
        \bottomrule
    \end{tabular}
    }
    \caption{Comparing to the existing VQA benchmarks \cite{goyal2017making, hudson2019gqa, johnson2017clevr, schwenk2022okvqa}, Streaming VQA is the first that provides a sequence of images for each query. }
\end{table*}

\paragraph{Annotation and Validation} A two-step validation process is employed. First, a multi-modal LLM, \texttt{GPT-4o-mini} \cite{openai20244o}, verifies each image by generating an answer and comparing it to the expected answer. Only images where the MLLM's response matches the assigned answer are retained. Then, human evaluators verify a random image subset, achieving approximately 93\% accuracy. The final benchmark encompasses five reasoning types:
\begin{itemize}[leftmargin=0.15in, itemsep=0em]
    \item \textbf{Spatial Reasoning} (13 queries): These questions require understanding the spatial relationships between objects within a scene, \eg “\textit{What is in the jar to the left of the juice?}”.
    \item \textbf{Logical Reasoning} (9 queries): This category involves applying conditions, rules, or filters to identify specific objects or answer complex queries, \eg “\textit{What is the black object on the desk that is not electronic?}”, “\textit{How many people are wearing both glasses and a hat?}”.
    \item \textbf{Numerical Reasoning} (11 queries): This category requires counting, comparing numbers, or calculating quantities based on visual information, \eg “\textit{How many extra bottles of beer do we need to make it a half dozen?}”.
    \item \textbf{Comparative Reasoning} (11 queries): These questions involve evaluating two or more objects in terms of their attributes, such as size, height, quantity, or quality, \eg “\textit{Which bottle is taller, the left one or the right one?}”.
    \item \textbf{External Knowledge Reasoning} (6 queries): These questions rely on information that extends beyond what is immediately visible in the image, often drawing on common sense or factual knowledge, \eg “\textit{The fruit in the picture is a good source of what vitamin?}”, “\textit{How many states are there in the country whose flag is being displayed?}”.
\end{itemize}
\paragraph{Evaluation} Performance for the streaming VQA task is evaluated using exact match accuracy, measuring the proportion of questions answered correctly without partial credit.
\end{document}